\documentclass[10pt,twocolumn,letterpaper]{article}

\usepackage[pagenumbers]{cvpr} 
\usepackage{graphicx}
\usepackage{amsmath}
\usepackage{amssymb}
\usepackage{booktabs}

\usepackage{multirow}
\usepackage{makecell}
\usepackage{subcaption}
\usepackage{url}

\usepackage[pagebackref,linkcolor=red,breaklinks,colorlinks]{hyperref}
\makeatletter
\newcommand{\printfnsymbol}[1]{%
  \textsuperscript{\@fnsymbol{#1}}%
}
\makeatother

\usepackage[capitalize]{cleveref}
\crefname{section}{Sec.}{Secs.}
\Crefname{section}{Section}{Sections}
\Crefname{table}{Table}{Tables}
\crefname{table}{Tab.}{Tabs.}

\def\cvprPaperID{2967} 
\def\confName{CVPR}
\def\confYear{2023}

\begin{document}

\title{Rethinking the Approximation Error in 3D Surface Fitting for Point Cloud Normal Estimation}

\author{{Hang Du\thanks{These authors contributed equally to this work.}, Xuejun Yan\printfnsymbol{1}, Jingjing Wang, Di Xie\thanks{Corresponding author.} , and Shiliang Pu}
\\
{Hikvision Research Institute, Hangzhou, China}
\\
\tt \small {\{duhang, yanxuejun, wangjingjing9, xiedi, pushiliang.hri\}@hikvision.com}
}
\maketitle

\begin{abstract}
Most existing approaches for point cloud normal estimation aim to locally fit a geometric surface and calculate the normal from the fitted surface. 
Recently, learning-based methods have adopted a routine of predicting point-wise weights to solve the weighted least-squares surface fitting problem.
Despite achieving remarkable progress, these methods overlook the approximation error of the fitting problem, resulting in a less accurate fitted surface. 
In this paper, we first carry out in-depth analysis of the approximation error in the surface fitting problem.
Then, in order to bridge the gap between estimated and precise surface normals, we present two basic design principles:
1) applies the $Z$-direction Transform to rotate local patches for a better surface fitting with a lower approximation error;
2) models the error of the normal estimation as a learnable term. 
We implement these two principles using deep neural networks, and integrate them with the state-of-the-art (SOTA) normal estimation methods in a plug-and-play manner. 
Extensive experiments verify our approaches bring benefits to point cloud normal estimation and push the frontier of state-of-the-art performance on both synthetic and real-world datasets. The code is available at~\url{https://github.com/hikvision-research/3DVision}. 

\end{abstract}

\section{Introduction}
\label{sec:intro}

Surface normal estimation on point clouds can offer additional local geometric information for numerous  applications, such as denoising~\cite{lu2020low,lu2020deep,edirimuni2022contrastive}, segmentation~\cite{qi2017pointnet,qi2017pointnet++,qian2022pointnext}, registration~\cite{pomerleau2015review,wang2019deep,huang2021predator,dai2023MDR-MFI}, and surface reconstruction~\cite{hoppe1992surface,kazhdan2006poisson,ma2021neural,du2022point,zhu2022semi}. 
However, raw-scanned point clouds tend to be incomplete, noisy, and non-uniform, which poses a challenge in accurately estimating surface normals amidst noise, density variations, and missing structures.

Normal estimation on point clouds is a long-standing research topic. The majority of traditional methods~\cite{hoppe1992surface,levin1998approximation,cazals2005estimating,guennebaud2007algebraic,boulch2012fast} aim to fit a local geometric surface (\eg, plane, jet and spherical) around a specific point, and infer the normal from the fitted surface. 
However, these methods require to carefully tune the setting of parameters, such as point neighborhood sizes, which is sensitive to noise and outliers.
With the power of deep neural networks, many learning-based approaches~\cite{guerrero2018PCPNet,ben20183dmfv,ben2019nesti,qian2020pugeo,zhou2020geometry,zhou2022fast,hashimoto2019normal} have been proposed to regress surface normal vectors directly, achieving promising performance improvements over traditional methods. 
However, these approaches exhibit limited generalization capability when applied to real-world point clouds.

\begin{figure}[t]
    \centering
    \includegraphics[height=7.4cm]{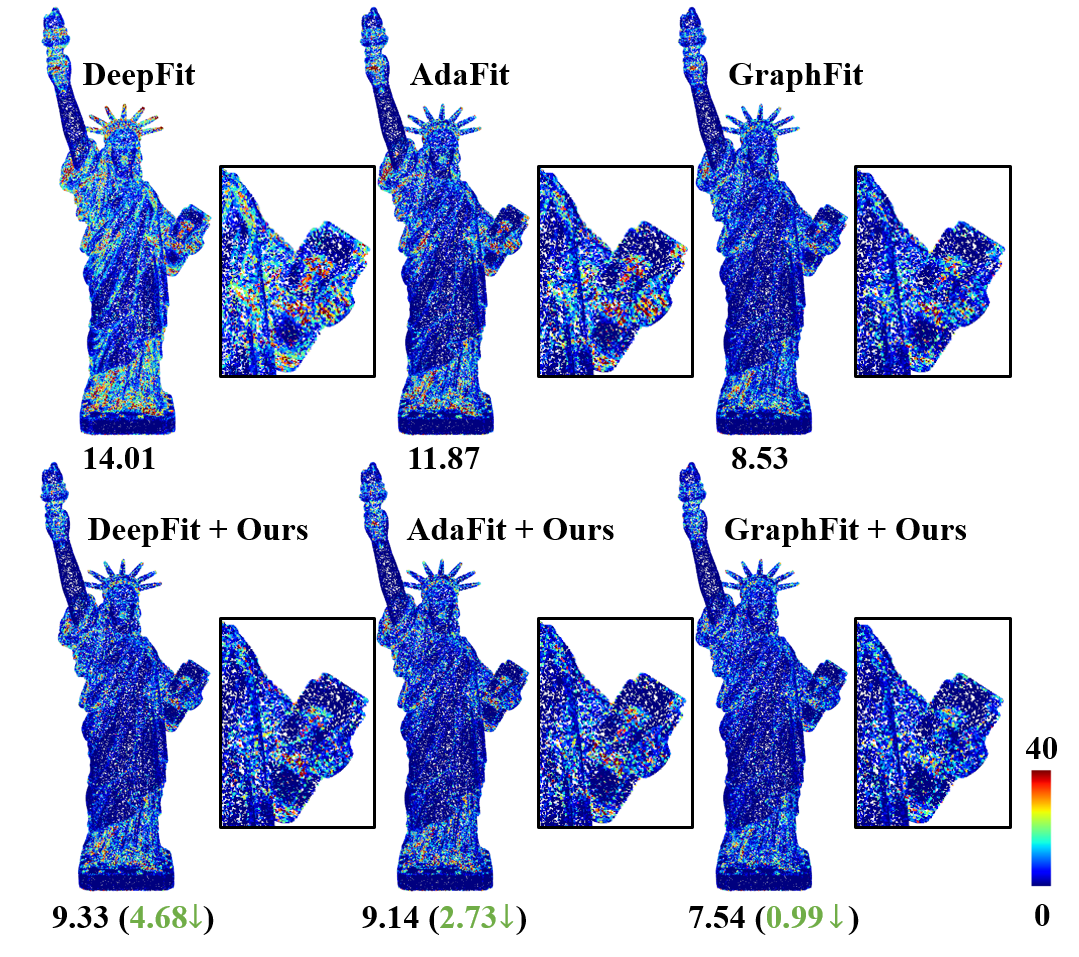}
    \caption{The error heatmap of point cloud normal estimation. The first row is produced by three SOTA surface fitting methods, while the second row shows the results of integrating our method with them. 
    The bottom values indicate the corresponding normal angle root mean square error (RMSE) . Both quantitative and qualitative results demonstrate that our proposed method provides more precise normal estimation.}
    \label{fig_1}
    \vspace{-1em}
\end{figure}

More recently, several approaches~\cite{ben2020deepfit,zhu2021adafit,li2022graphfit} have generalized the truncated Taylor expansion ($n$-jet) surface model~\cite{cazals2005estimating} to the learning-based regime, formulating normal estimation as a weighted least-squares problem with learnable weights. 
In these methods, the point-wise weights of a local surface patch are predicted by a deep neural network, which can control the importance of neighboring points to the fitted surface and alleviate the sensitivity to outliers and noise. 
Then, the solution of weighted least-squares fitting problem can be expressed in a closed form, which enables to estimate the geometric surface and infer the surface normal. 
These methods heavily constrain the solution space and obtain a better result for surface normal estimation.
Nevertheless, none of them theoretically analyzes the approximation error in surface fitting, leading to a suboptimal normal estimation performance. 
In some sense, a smaller approximation error represents a more precise estimation. 
Therefore, we aim to study how to reduce the approximation error and fit a more accurate surface for normal estimation. 

In this paper, we analyze the approximation error in the $n$-jet surface model, and find the existing gap between estimated and accurate normals in previous methods. 
Specifically, the truncated Taylor expansion polynomial is expected to be equivalent to the height function of the surface, and the accuracy of the reminder term in Taylor expansion has an impact on the precision of normal estimation. 
As pointed out in~\cite{cazals2005estimating}, to improve the accuracy, a feasible way is to set up a coordinate system where $z$ direction is aligned (has the minimum angle) with the estimated normal. 
However, we find the previous methods cannot accomplish this objective well, leading to a large estimation error in most cases. 
Besides, due to the presence of the reminder term and the imperfect data (inevitably containing outliers and noise), it is impossible to achieve an accurate surface fitting without any approximation error. 
To solve these problems, we propose two basic design principles. 
First, we apply the $z$-direction transformation to rotate local patches for a better surface fitting with a lower approximation error. 
Second, the error of normal estimation is modeled as a term that can be learned in a data-driven manner. 
The proposed principles can improve the accuracy of the surface fitting, thereby leading to a more precise estimation of surface normals.

To model the above two principles, we implement them with deep neural networks, and introduce two simple yet effective methods: $Z$-direction Transformation and Normal Error Estimation.  
More specifically, the $z$-direction transformation is fulfilled by adding a constraint on the angle between the rotated normal and the $z$ axis, which aims to align the rotated normal with the $z$ direction.
To achieve this learning objective, we also design a graph-convolution based alignment transformation network to fully exploit the local neighborhood information for learning a better point transformation. 
Then, the rough estimated normal can be inferred by any existing polynomial surface fitting method, such as DeepFit~\cite{ben2020deepfit} and GraphFit~\cite{li2022graphfit}. 
Finally, we design a normal error estimation module that learns a residual term based on the rough estimated result and thus improves the precision of normal estimation. 

We conduct comprehensive experiments to verify the effectiveness of our methods on point cloud normal estimation. 
The proposed two basic design principles are implemented with the existing polynomial surface fitting methods. 
The experimental results demonstrate our design principles are beneficial to these methods with few extra burdens. 
As shown in Fig.~\ref{fig_1}, an obvious improvement can be achieved by our proposed methods for normal estimation. 

The contributions of this paper are summarized as: 
\begin{itemize}

\item We provide an in-depth analysis of the approximation error in $n$-jet surface fitting, and introduce two basic design principles to improve the precision of 3D surface fitting.

\item We implement the design principles with neural networks and propose two approaches, \ie, $z$-direction transformation and normal error estimation, which can be flexibly integrated with the current polynomial surface fitting methods for point cloud normal estimation. 

\item We conduct extensive experiments to show the improvements by the proposed methods. The experimental results demonstrate our methods consistently bring benefits and push the frontier of SOTA performance.

\end{itemize}

\section{Related Work}
\label{sec:related_work}
\subsection{Traditional Approaches}
Normal Estimation on point clouds has been widely studied. A commonly-used way is the Principal Component Analysis (PCA), which can be utilized to estimate a tangent plane by computing the eigenvector with the smallest eigenvalue of a covariance matrix~\cite{hoppe1992surface}. 
Subsequently, some approaches~\cite{levin1998approximation,guennebaud2007algebraic,cazals2005estimating} are designed to fit a more complex surface (\eg, jet and spherical) by involving more neighboring points. 
Although these methods enable to be more robust to the noise and outliers, the shape details are over-smoothed due to the large neighborhood size. 
In order to preserve the shape details, certain methods employ Voronoi diagram~\cite{alliez2007voronoi,amenta1998surface,merigot2010voronoi} or Hough transform~\cite{boulch2012fast} for normal estimation. 
However, they require careful parameters tuning to handle the input points with different noise levels. 
The above-mentioned methods are sensitive to the setting of parameters, such as the point neighborhood sizes.
There is no universal setting that can meet all the challenges.

\subsection{Learning-based Methods}
Learning-based methods~\cite{boulch2016deep,guerrero2018PCPNet,ben2019nesti,lenssen2020deep,ben2020deepfit,lu2020deep,zhou2020normal,zhu2021adafit,li2022graphfit,wang2022deep,zhang2022geometry,li2022hsurf} have better robustness for noise and outliers, which can be roughly divided into regression and surface fitting based methods.

\noindent\textbf{Regression based}. The estimation of surface normals can be regressed by deep neural networks directly. A group of methods~\cite{boulch2016deep,zhou2020geometry,lu2020deep} aim to transform the input points into structured data, such as 2D grid representations, and train a Convolutional Neural Network (CNN) to predict the normal vectors. Another kind of methods~\cite{ben20183dmfv,guerrero2018PCPNet,hashimoto2019normal,zhou2020normal,qian2020pugeo,zhou2022refine,li2022hsurf} takes the advantages of point cloud processing network and directly predicts surface normals from unstructured point clouds. 
For example, PCPNet~\cite{guerrero2018PCPNet} adopts a multi-scale PointNet~\cite{qi2017pointnet} to process the point clouds with different neighborhood sizes simultaneously, and the patch-wise global features are regressed to the geometric properties consequently. Moreover, Nesti-Net~\cite{ben2019nesti} and Zhou~\etal~\cite{zhou2020normal} propose to select the optimal neighborhood size by designing a multi-scale architecture and choose the corresponding sub-network for normal estimation. Although these methods perform better than traditional routine, they lack of generalization ability on unseen data.

\noindent\textbf{Surface fitting based}. In contrast to regressing the normal vectors directly, recent approaches~\cite{lenssen2020deep,ben2020deepfit,zhu2021adafit,li2022graphfit} aim to integrate deep learning with solving the least-squares surface fitting problem. Among them, lenssen~\etal~\cite{lenssen2020deep} propose an iterative estimation method that predicts point weights for plane fitting and iteratively adjusts the weights to a better fitting. 
Other methods~\cite{ben2020deepfit,zhu2021adafit,li2022graphfit} extend the $n$-jet surface fitting~\cite{cazals2005estimating} to the learning-based regime. 
They also employ the neural network to predict point-wise weights which can be regarded as a selection of sample points. 
The surface normal is calculated from the solution of polynomial surface fitting problem. 
To improve the precision of normal estimation, AdaFit~\cite{zhu2021adafit} adjusts the position of neighboring points by regressing the point offsets, and GraphFit~\cite{li2022graphfit} designs a more powerful network for learning point-wise weights. 
In this work, we follow the same routine and study the approximation error of polynomial surface fitting methods.
To bridge the gap between estimated and precise surface normals, we propose two basic design principles which can be integrated with the current polynomial fitting methods and improve their performance on surface normal estimation.

\section{Theoretical Formulation}

The truncated Taylor expansion ($n$-jet) surface model has been widely used for estimating geometric quantities, such as normal vectors, and curvatures. In this section, we first revisit the theory of $n$-jet surface model, and then analyze the approximation error of $n$-jet fitting problem, which facilitates the next section to present our two basic design principles that pursue a more precise normal estimation.

\subsection{Revisiting $N$-jet Surface Fitting}

Jet surface model~\cite{cazals2005estimating} represents a polynomial function that mapping points $(x, y)$ of $\mathbb{R}^2$ to their height $z \in \mathbb{R}$ over a surface, where any $z$ axis is not in the tangent space. In other words, given a point $(x, y)$, $z$ on the surface can be obtained by the height function $f(x, y)$. Then, an order $n$ Taylor expansion of the height function over a surface is defined as: 
\begin{equation}
\label{n_taylor}
f(x, y)=J_{\beta, n}(x, y)+\mathrm{O}\left(\|(x, y)\|^{n+1}\right), 
\end{equation} 
where the truncated Taylor expansion $J_{\beta, n}(x, y)=\sum_{k=0}^n \sum_{j=0}^k \beta_{k-j, j} x^{k-j} y^j$ is called a degree $n$ jet, or $n$-jet, and $\mathrm{O}(\|(x, y)\|^{n+1})$ denotes the remainder term of Taylor expansion. 
Then, the surface normal given by Eq.~\ref{n_taylor} is 
\begin{equation}
\label{normal_es}
\hat{\mathbf{n}}=\frac{\left(-\beta_{1,0},-\beta_{0,1},1\right)}
{\sqrt{\beta_{1,0}^2+\beta_{0,1}^2+ 1}}.
\end{equation}

In the approximation case, the number of point $N_p$ is larger than that of the  coefficients $N_n= {(n+1)(n+2) }/{2}$. The $n$-jet surface model aims to find an approximation result on the coefficients of the height function, which can be expressed as:  
\vspace{-0.5em}
\begin{equation}
\label{fitting_problem}
J_{\alpha, n}=\arg \min \left\{\sum_{i=1}^{N_p}\left(J_{\alpha, n}\left(x_i, y_i\right)-f\left(x_i, y_i\right)\right)^2\right\}, 
\end{equation}
where $J_{\alpha, n}$ is the solution of the least-square polynomial fitting problem, and $i=1,...,N_p$ is the index of a set of points. 
Moreover, considering the noise and outliers have a large impact on the fitting accuracy, a widely-used way is to extend Eq.~\ref{fitting_problem} to a weighted least-square problem, and thus the solution can be expressed as: 
\begin{equation}
\label{solution}
{\boldsymbol{\alpha}}=\left(\mathbf{M}^{\top} \mathbf{W} \mathbf{ M}\right)^{-1}\left(\mathbf{M}^{\top} \mathbf{W}  \mathbf{z}\right), 
\end{equation} 
where $\mathbf{M}=\left(1, x_i, y_i, \ldots, x_i y_i^{n-1}, y_i^n\right)_{i=1, \ldots, N_p} \in \mathbb{R}^{N_p \times N_n}$ is the Vandermonde matrix, $\mathbf{W} =\operatorname{diag}\left(w_1, w_2, \ldots, w_{N_p}\right) \in \mathbb{R}^{N_p \times N_p}$ is the diagonal point-wise weight matrix and 
$\mathbf{z}=\left(z_1, z_2, \ldots, z_{N_p}\right) \in \mathbb{R}^{N_p}$ is the ${N_p}$-vector of coordinates.

\begin{figure*}[ht]
    \centering
    \includegraphics[height=6.7cm]{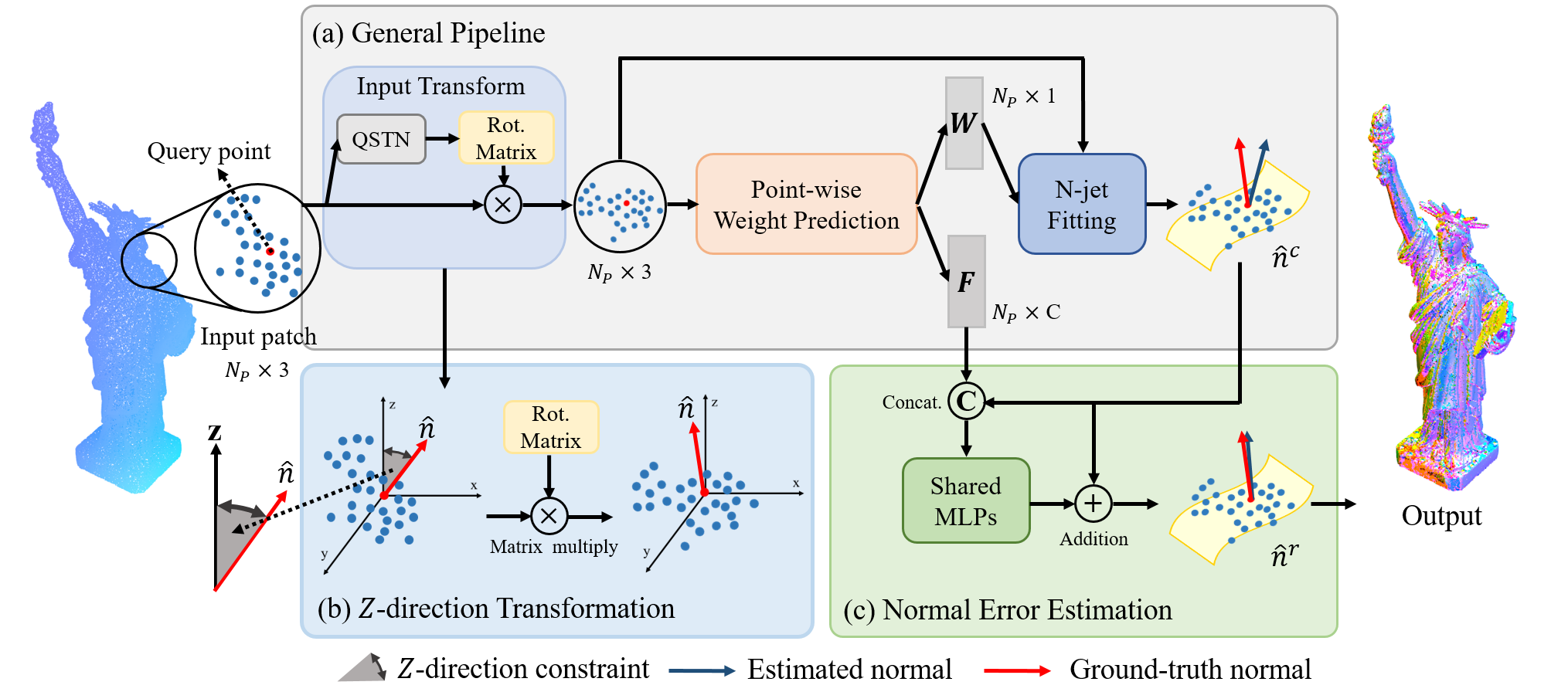}
    \caption{(a) The overall architecture of a general $n$-jet fitting network. The input patch is fed into the spatial transformation network to learn a quaternion for rotation transformation. Then, point-wise weights are predicted from the rotated points. Finally, the weight predictions are utilized to solve $n$-jet fitting problem, and the normal is inferred from the fitted surface.   
    (b) The proposed \textbf{$Z$-direction Transformation}: the rotation matrix is explicitly constrained by $z$-direction transformation, which aims to align the rotated normal to the $z$ axis.
    (c) The proposed \textbf{Normal Error Estimation}: the rough estimated surface normal is further adjusted by learnable error estimation.}
    \label{framework}
    \vspace{-1em}
\end{figure*}

\subsection{Analysis of Approximation Error}
\label{analysis_approximation}

Recent surface fitting methods~\cite{ben2020deepfit,zhu2021adafit,li2022graphfit} have generalized the $n$-jet surface model to the learning-based regime, which predict the point-wise weights resorting to the deep neural network. 
They obtain the coefficients of $J_{\alpha, n}$ by solving the weighted least-squares polynomial fitting problem, and expect that the coefficients of $J_{\alpha, n}$ are approximated to those of $J_{\beta, n}$. 
In this way, the surface normal can be calculated by Eq.~\ref{normal_es}. 

However, as presented in~\cite{cazals2005estimating}, assuming the convergence rate of approximation is given by the value of the exponent of parameter $h$ and $\mathrm{O} (h) = \|(x_i, y_i)\|$, the coefficients $\beta_{k-j, j}$ of $J_{\beta, n}(x, y)$ are estimated by those of $J_{\alpha, n}(x, y)$ up to accuracy $\mathrm{O}(h^{n-k+1})$: 
\begin{equation}
\label{first_coefficients}
\alpha_{k-j, j}=\beta_{k-j, j}+\mathrm{O}(h^{n-k+1}).  
\end{equation}
Hence, there exists a error term $\mathrm{O}(h^{n-k+1})$ between the coefficients $\beta_{k-j, j}$ of $J_{\beta, n}$ and those of $J_{\alpha, n}$, 
and a smaller error enables to yield a more precise  estimation result. 

As proof in~\cite{cazals2005estimating},~\textit{the error estimates are better when the convex hull of the sample points is not too  ``flat''}. To be specific,  $\mathrm{O}\left(h^{n-k+1}\right)$ depends on the supremum of $\left\{\left\|D^{n+1}f(x, y)\right\| ; (x, y) \in K \right\}$, where $K$ is the convex hull of the point set $\left\{\left(x_i, y_i\right)\right\}_{i=1, \ldots, N}$. 
Let $d_{max}$ be the diameter of $K$ and $d_{min}$ be the supremum of the diameter of disks inscribed in $K$. 
A small ratio $d_{max}/d_{min}$ means that the geometry of $K$ is not too ``flat'', which can reduce the supremum of $\left\|D^{n+1}f(x, y)\right\|$, leading to a better surface fitting with a lower error. 
To decrease the ratio, one should take a coordinate system as close as possible to the Monge system, where the $z$ axis is aligned with the true normal. 
So, a good choice is to rotate the point cloud and minimize the angle between the rotated normal and the $z$ axis. 
By doing so, one can reduce the approximation error and confirm the convergence of the surface fitting problem well. 

In addition, \textit{a polynomial fitting of degree $n$ estimates any $k$th-order differential quantity to accuracy $\mathrm{O}(h^{n-k+1})$}. 
Here, we take the notations of a curve for simplicity, 
\begin{equation}
 \label{angle_error}
\small
\begin{aligned}
 &F\left(\alpha_{i=0, \ldots, k}\right)
  = F\left(\beta_{i=0, \ldots, k} + \mathrm{O}(h^{n-k+1})\right) 
 \\
 &=F\left(\beta_{i=0, \ldots, k}\right)+DF_{(\beta_i+ \mathrm{O}(h^{n-k+1}))_{i=0, \ldots, k}}\left(\mathrm{O}(h^{n-k+1})\right),
 \end{aligned}
\end{equation}
where $F(\alpha_{i})$ is the differential quantity (\eg, curvatures or normals) function of the $n$-jet coefficients $\alpha_{i=0, \ldots, k}$, and $DF_{p}$ is the differential of $F$ at point $p$. 
For normal estimation, \ie, $k=1$, the error term $DF_{\left(\beta_i+ \mathrm{O}(h^{n})\right)}\left(\mathrm{O}(h^{n})\right)$ denotes the angle between the true normal and the estimated normal. 
Due to the presence of this error term and imperfect data, obtaining precise estimations remains a challenge.  
Thereby, the normal error estimation is another good choice to improve the precision of surface normal estimation.

\section{The Proposed Approach}

Motivated by the theoretical formulation outlined above, we introduce two basic design principles to reduce the approximation error and improve the precision of surface normal estimation: 
1) we aim to explicitly learn a alignment transformation that rotates input patches for a better surface fitting with a lower approximation error; 
2) we model the error of normal estimation as a learnable term which compensates the rough estimated surface normal and yields a more accurate result. 
Based on these two basic design principles, we propose two simple yet effective methods, \ie, $z$-direction transformation and normal error estimation, which can be flexibly integrated with the current polynomial surface fitting model for further improvements.  
In the following, we first provide the overview of $n$-jet surface normal estimation network, and then elaborate the implementation details of the proposed methods.

\subsection{Overview}

As shown in Fig.~\ref{framework}, a general $n$-jet surface fitting network consists of three components, including input transformation, point-wise weight prediction, and normal estimation ($n$-jet fitting). 
For input transformation, we propose to explicitly constrain the learned transformation matrix, which aims to narrow the angle between the rotated normal and the $z$ axis. 
Then, the point-wise weights prediction module can be any existing polynomial surface fitting network, such as DeepFit~\cite{ben2020deepfit}, AdaFit~\cite{zhu2021adafit}, and GraphFit~\cite{li2022graphfit}.
The weights are utilized for solving the weighted least-square polynomial fitting problem in Eq.~\ref{solution}, and the rough normal can be calculated from the fitted surface. 
Finally, we learn a normal error term and add it on the rough estimated result to yield a more precise estimation.

\begin{figure}[t]
\begin{minipage}[t]{0.45\linewidth}
\includegraphics[height=3.1cm]{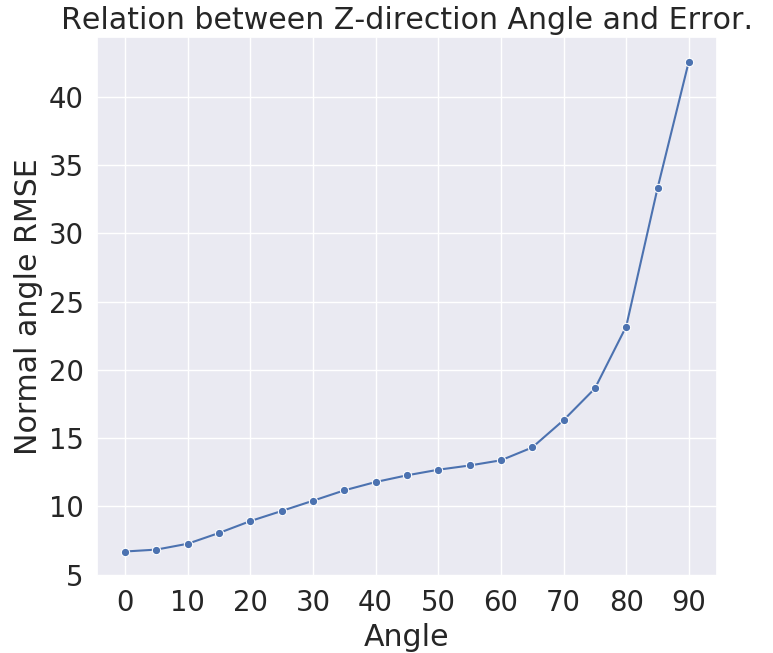}
\subcaption{}
\label{error_bar}
\end{minipage} 
 \hfill
\begin{minipage}[t]{0.54\linewidth}
\includegraphics[height=3.1cm]{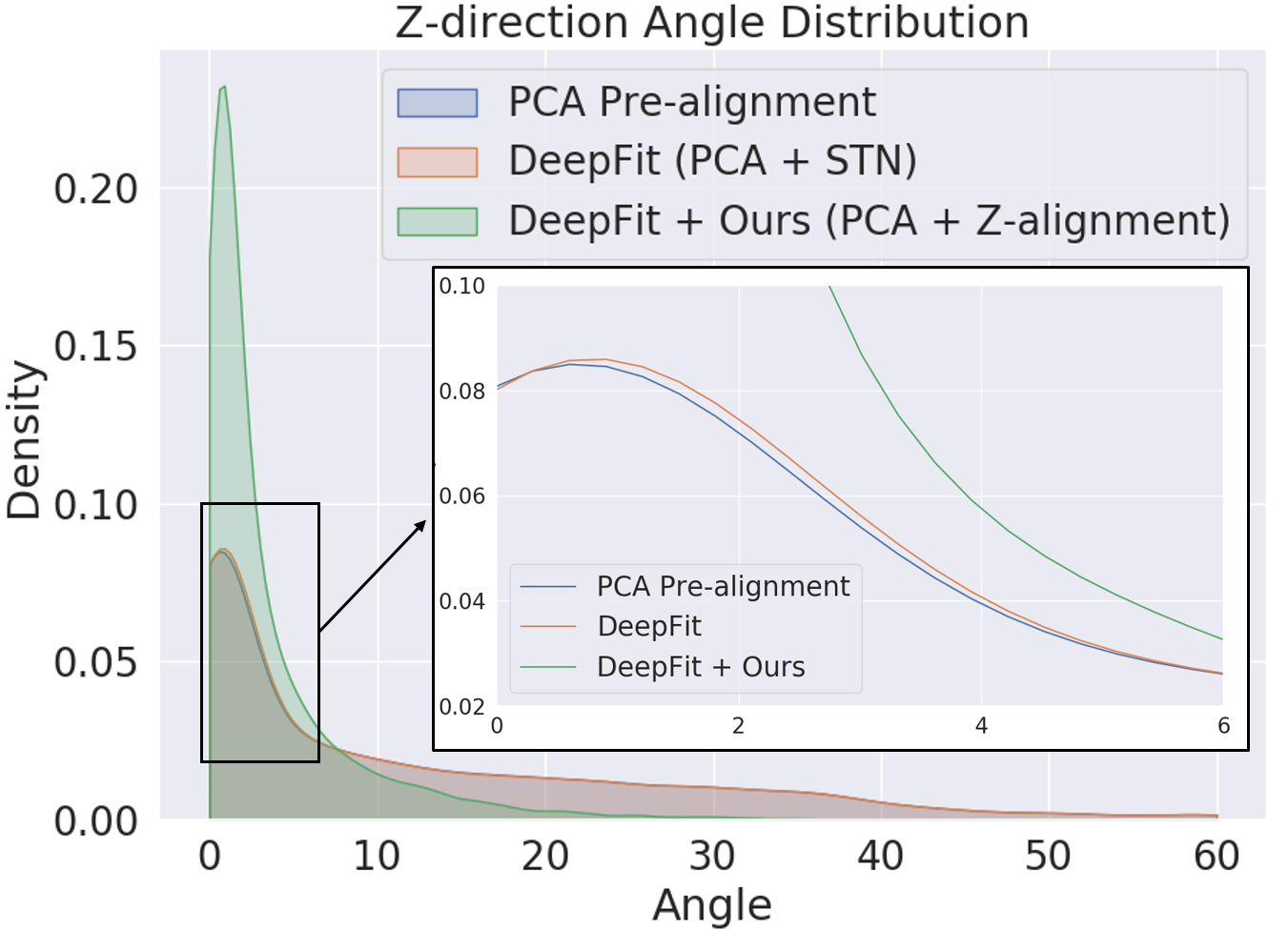}
\centering
\subcaption{}
\label{angle_dis}
\end{minipage} 
\vspace{-1em}
\caption{The analysis of $z$-direction transformation.
(a) The relation between the error of normal estimation and the $z$-direction angle. A larger $z$-direction angle usually increases the error. 
(b) The distribution of the angle between the rotated true normal and the $z$ axis. In the original DeepFit~\cite{ben2020deepfit}, approximate $30\%$ cases exhibit a large angle ($\geq 20$) with the $z$ axis after transformation.
}
\vspace{-1.5em}
\end{figure}

\subsection{$Z$-direction Transformation} 
In order to learn transformation invariant features, PointNet~\cite{qi2017pointnet} adopts a spatial transformation network (STN) to align the input points to a canonical view. 
In terms of surface normal estimation, rotation transformation is a more proper choice, since it can stabilize the convergence of network~\cite{guerrero2018PCPNet}. Thus, previous methods~\cite{guerrero2018PCPNet,ben2020deepfit,zhu2021adafit,li2022graphfit} tend to learn a quaternion for rotation transformation. 

As explained in Sec.~\ref{analysis_approximation}, the accuracy of surface fitting problem can be improved by minimizing the angle between the rotated normal and the $z$ axis. 
Here, we argue that the error of normal estimation can be decreased when the $z$-direction angle is minimized. 
To further support this argument, we conduct a toy experiment on PCPNet~\cite{guerrero2018PCPNet} test set, and analyze the relation between the $z$-direction angle and the normal estimation error. 
In Fig.~\ref{error_bar}, we can find the error is positively related to the $z$-direction angle of rotated normals, which means a larger $z$-direction angle usually increases the normal error. 
Moreover, we inspect the angle distribution between the rotated ground-truth normal and the $z$ axis. 
The angle distribution is shown in Fig.~\ref{angle_dis}, where the horizontal axis denotes the value of angle and the vertical axis denotes the density.
In the original DeepFit model, approximate $30\%$ cases exhibit a large angle ($\geq20$) with the $z$ axis after transformation suggesting that the model is unable to learn the optimal transformation for achieving the best surface fitting, Thereby, it fails to obtain a precise normal estimation result.

From the above analysis, we can see the mechanism of $z$-direction alignment transformation enables to reduce the approximation error and thus improve the accuracy of the surface fitting problem. 
However, the existing transformation network cannot fulfill this objective well. 
To address it, we aim to explicitly learn a quaternion spatial transformation $\mathbf{T}$ that sets the rotated ground-truth normal aligned with the axis $z$, which can be formulated as 
\begin{equation}
\hat{\mathbf{z}} =\hat{\mathbf{n}}_{i} \mathbf{T}, 
\end{equation}
where $\hat{\mathbf{z}} = (0,0,1)^{\top}$, $\hat{ \mathbf{n}}_i$ is the ground-truth normal, and $\mathbf{T}$ is the learned transformation. We expect the rotated normal has the minimum angle with the $z$ axis. To accomplish this objective, we design a $z$-direction transformation loss, 
\begin{equation}
\label{z_loss}
\mathcal{L}_\text{trans} =  \left|\hat{\mathbf{n}}_i \mathbf{T} \times \hat{\mathbf{z}} \right|. 
\end{equation}
This loss term can explicitly constrain the learned alignment transformation towards the $z$ direction.
However, the expected transformation is non-trivial by simply using a chain of Multi-layer Perceptrons (MLPs) in~\cite{guerrero2018PCPNet,ben2020deepfit,zhu2021adafit}. 
To this end, we design a graph-convolution (GCN) based spatial  transformation network by fully exploiting the local neighborhood information. 
Specifically, three EdgeConv~\cite{wang2019dynamic} layers with two adaptive graph pooling~\cite{yan2022fbnet} are utilized to learn multi-scale point features. 
Then, a combination of max and average pooling operations is adopted to generate a global feature which is further reduced to a quaternion for $z$-direction transformation. 
As shown in Fig.~\ref{angle_dis}, more angles tend to zero under the green curve, which means our $z$-direction transformation indeed aligns the normal to the $z$ axis, and thereby leads to a more accurate surface fitting.  

\begin{table*}[t]
    \begin{center}
    \caption{Normal angle RMSE of our methods and baseline models on PCPNet dataset. After being integrated with our methods, the state-of-the-art surface fitting methods obtain significant improvements on point cloud normal estimation. }
    \label{sota_performance}
    \resizebox{1\linewidth}{!}{
    \begin{tabular}{c|ccc|cccccccc}
    \toprule[1pt]
  {\textbf{Aug.}}& \makecell[c]{GrapFit\\ + Ours}& \makecell[c]{AdaFit\\ + Ours}& \makecell[c]{DeepFit \\ + Ours}& \makecell[c]{GraphFit \\ ~\cite{li2022graphfit}} &\makecell[c]{AdaFit \\ \cite{zhu2021adafit}} &\makecell[c]{DeepFit\\ \cite{ben2020deepfit}} &\makecell[c]{IterNet \\ \cite{lenssen2020deep}}&\makecell[c]{Nesti-Net \\ \cite{ben2019nesti}}&\makecell[c]{PCPNet \\ \cite{guerrero2018PCPNet}}&\makecell[c]{Jet\\ \cite{cazals2005estimating}}&\makecell[c]{PCA \\ \cite{hoppe1992surface}}\\
    \midrule[0.5pt]
  No Noise &~\textbf{4.11}&4.71&4.90&4.45&5.19&6.51&6.72&6.99&9.62&12.25&12.29\\
  Noise ($\sigma=0.125\%$) &~\textbf{8.66}&8.75&8.91&8.74&9.05&9.21&9.95&10.11&11.37&12.84&12.87\\
  Noise ($\sigma=0.6\%$) &~\textbf{16.02}&16.31&16.61&16.05&16.44&16.72&17.18&17.63&18.87&18.33&18.38\\
  Noise ($\sigma=1.2\%$) &~\textbf{21.57}&21.64&22.87&21.64&21.94&23.12&21.96&22.28&23.28&27.68&27.50\\
  Density (Gradient) &~\textbf{4.83}&5.51&5.52&5.22&5.90&7.31&7.73&9.00&11.70&13.13&12.81\\
  Density (Striped) &~\textbf{4.89}&5.48&5.70&5.48&6.01&7.92&7.51&8.47&11.16&13.39&13.66\\
  \midrule[0.5pt]
  Average & \makecell[c]{~\textbf{10.01} \\ (~\textcolor{green}{\textbf{0.25}$\downarrow$})} & \makecell[c]{10.40 \\  (~\textcolor{green}{\textbf{0.36}$\downarrow$}) }&
  \makecell[c]{ 10.75 \\ (~\textcolor{green}{\textbf{1.05}$\downarrow$})} &10.26&10.76&11.80&11.84&12.41&14.34&16.29&16.25\\
    \bottomrule[1pt]
    \end{tabular}}
    \end{center}
    \vspace{-2em}
\end{table*} 

\subsection{Normal Error Estimation}
Although we can reduce the approximation error via $z$-alignment transformation, it is still challenging to yield precise normal estimations, due to the presence of error term $DF_{\left(\beta_i+ \mathrm{O}(h^{n})\right)}\left(\mathrm{O}(h^{n})\right)$ in the differential quantity function (Eq.~\ref{angle_error}) and imperfect data which inevitably contains the noise and outliers.
In the pursuit of a more precise normal estimation, we propose to estimate the error of normal estimation in a data-driven manner.  
To be specific, we consider the rough estimated normal $\hat{\mathbf{n}}_i^c $ of the fitted surface should be updated by learning a residual term, 
\begin{equation}
 \Delta(\hat{\mathbf{n}}_i^c) = \phi (\text{Concat}(x_i, \hat{\mathbf{n}}_i^c )), 
\end{equation}
where $x_i$ is the point-wise feature, and $\phi (\cdot)$ is a mapping function, \ie, MLPs. 
Then, we compute the final output normal $\hat{\mathbf{n}}_i^r$ by adding  $\Delta(\hat{\mathbf{n}}_i^c )$ on the rough estimated normal,  
\begin{equation}
\hat{\mathbf{n}}_i^r= \hat{\mathbf{n}}_i^c + \Delta(\hat{\mathbf{n}}_i^c).
\end{equation} 
In this way, the network also enables to adjust the inaccurate surface fitting brought by noise and outliers.
Finally, we can reduce the error of surface normal estimation and thereby yield a more accurate estimation result.

Note that the proposed normal error estimation is parallel to the point offset learning in AdaFit~\cite{zhu2021adafit}. 
In the following experiments, we can achieve a further improvement on AdaFit with our methods.
Moreover, compared with predicting the normal directly, residual error prediction is much easier and stable for the network.

\subsection{Implementation Details}
\noindent\textbf{Network Architecture.} In this study, we propose two basic design principles and implement them with DeepFit~\cite{ben2020deepfit}, AdaFit~\cite{zhu2021adafit}, and GraphFit~\cite{li2022graphfit}. So, we adopt their original network architecture with two displacements. 
First, the spatial transformation network is replaced with our GCN-based transformation network, and the learned transformation matrix is constrained by Eq.\ref{z_loss}. 
Second, we add two layers of MLPs to regress the residual terms for adjusting the rough estimated surface normal. 
More details on network architecture can be found in the supplementary materials. 

\noindent\textbf{Training Supervision.} 
To train the network, we employ the same loss functions in DeepFit~\cite{ben2020deepfit} for both rough estimated normal $\hat{\mathbf{n}}_i ^c$ and refined normal $\hat{\mathbf{n}}_i^r$,
\begin{equation}
\mathcal{L}_\text{normal} = \left|\hat{\mathbf{n}}_i  \times \hat{\mathbf{n}}_i ^c\right| + \left|\hat{\mathbf{n}}_i  \times \hat{\mathbf{n}}_i ^r\right|. 
\end{equation}
We also utilize the neighborhood consistency loss $\mathcal{L}_\text{con}$ and transformation regularization loss $\mathcal{L}_\text{reg}$ in DeepFit. 
Moreover, as above presented, we add a penalty term on transformation matrix $\mathcal{L}_\text{trans}$. Thus, the total training loss is
\begin{equation}
\mathcal{L}_\text{total} = \mathcal{L}_\text{normal} + \lambda_1 \mathcal{L}_\text{con} +  \lambda_2\mathcal{L}_\text{reg} +  \lambda_3\mathcal{L}_\text{trans},
\end{equation}
where we empirically set $\lambda_{1}=0.25,\lambda_{2}=0.1$, and $\lambda_{3}=2$ in the experiments.

\section{Experiment}

\subsection{Datasets and Experimental Settings} 

\noindent\textbf{Datasets.} We follow the same configuration of previous works that adopt synthetic PCPNet dataset for training, which includes four CAD objects and four high quality scans of figurines with total 3.2M training examples. Then, the trained models are evaluated on PCPNet test set with six categories, including four sets with different levels of noise, \ie, no noise, low noise ($\sigma=0.125\%$), med noise ($\sigma=0.6\%$), and high noise ($\sigma=1.2\%$), and two sets with varying sampling density (gradient and striped
pattern).
To verify the generalization ability, we also employ a real-world dataset, SceneNN~\cite{scenenn-3dv16}, for both quantitative and qualitative evaluation. 

\noindent\textbf{Training.} The polynomial order $n$ for the surface fitting is 3. Adam algorithm is used for model optimization. Our models are trained for 700 epochs with a batch size of 256. The learning rate begins at 0.001 and drops by a decay rate of 0.1 at 300 and 550 epochs.  

\noindent\textbf{Evaluation.} We select three recent state-of-the-art methods as baseline, including DeepFit~\cite{ben2020deepfit}, AdaFit~\cite{zhu2021adafit} and GraphFit~\cite{li2022graphfit}. 
Besides, we also compare with traditional methods and learning-based normal regression methods. 
The root-mean-squared error (RMSE) of angles between the estimated and the ground-truth normals to evaluate the performance. 
Moreover, we also report the percentage of good points with a threshold of error (PGP $\alpha$).

\subsection{SOTA Results on Synthetic Data}  
\begin{figure}[t]
\begin{minipage}[t]{0.47\linewidth}
\includegraphics[height=3.cm]{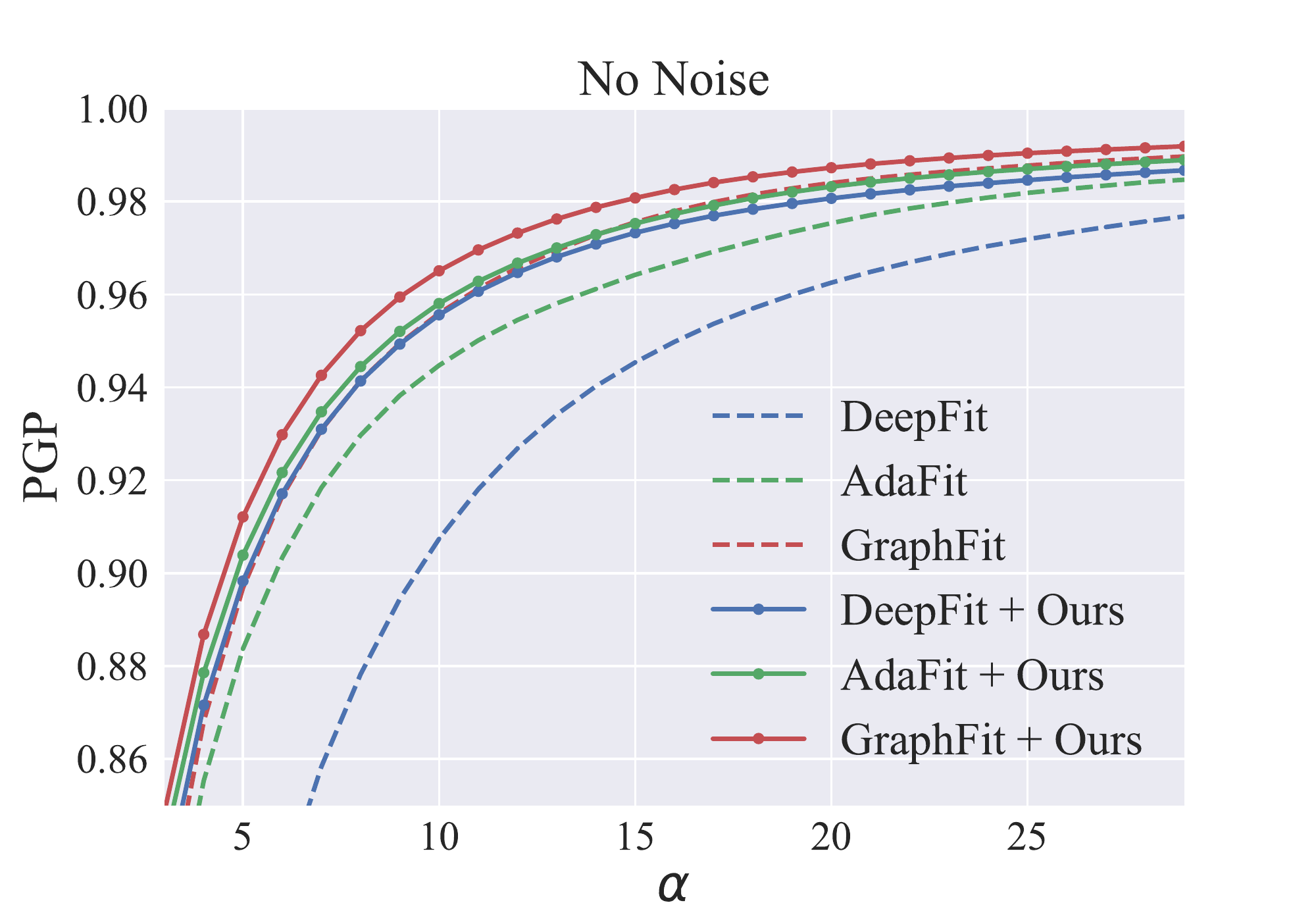}
\end{minipage} 
\begin{minipage}[t]{0.44\linewidth}
\includegraphics[height=3.cm]{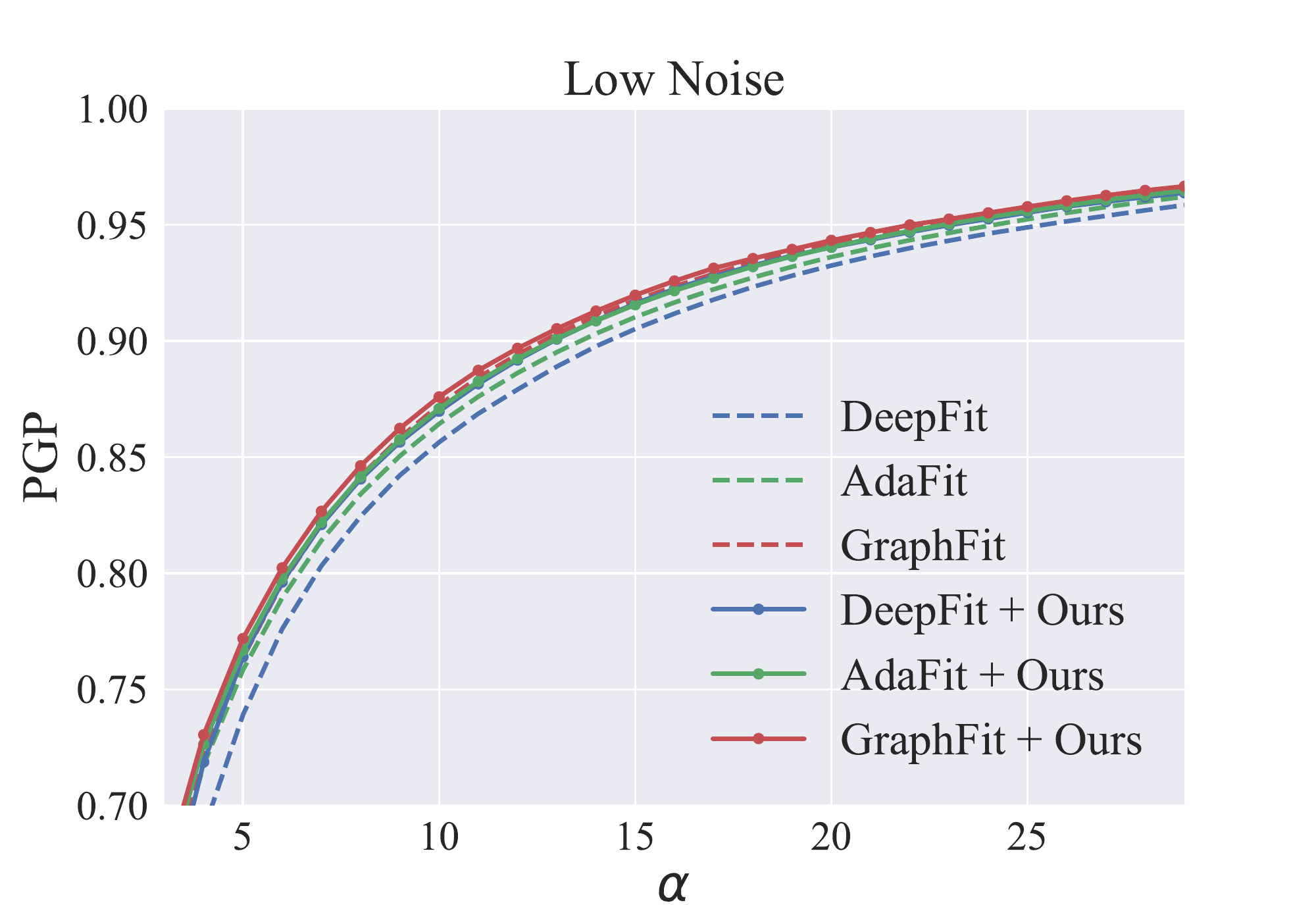}
\end{minipage} 
\\
\begin{minipage}[t]{0.47\linewidth}
\includegraphics[height=3.cm]{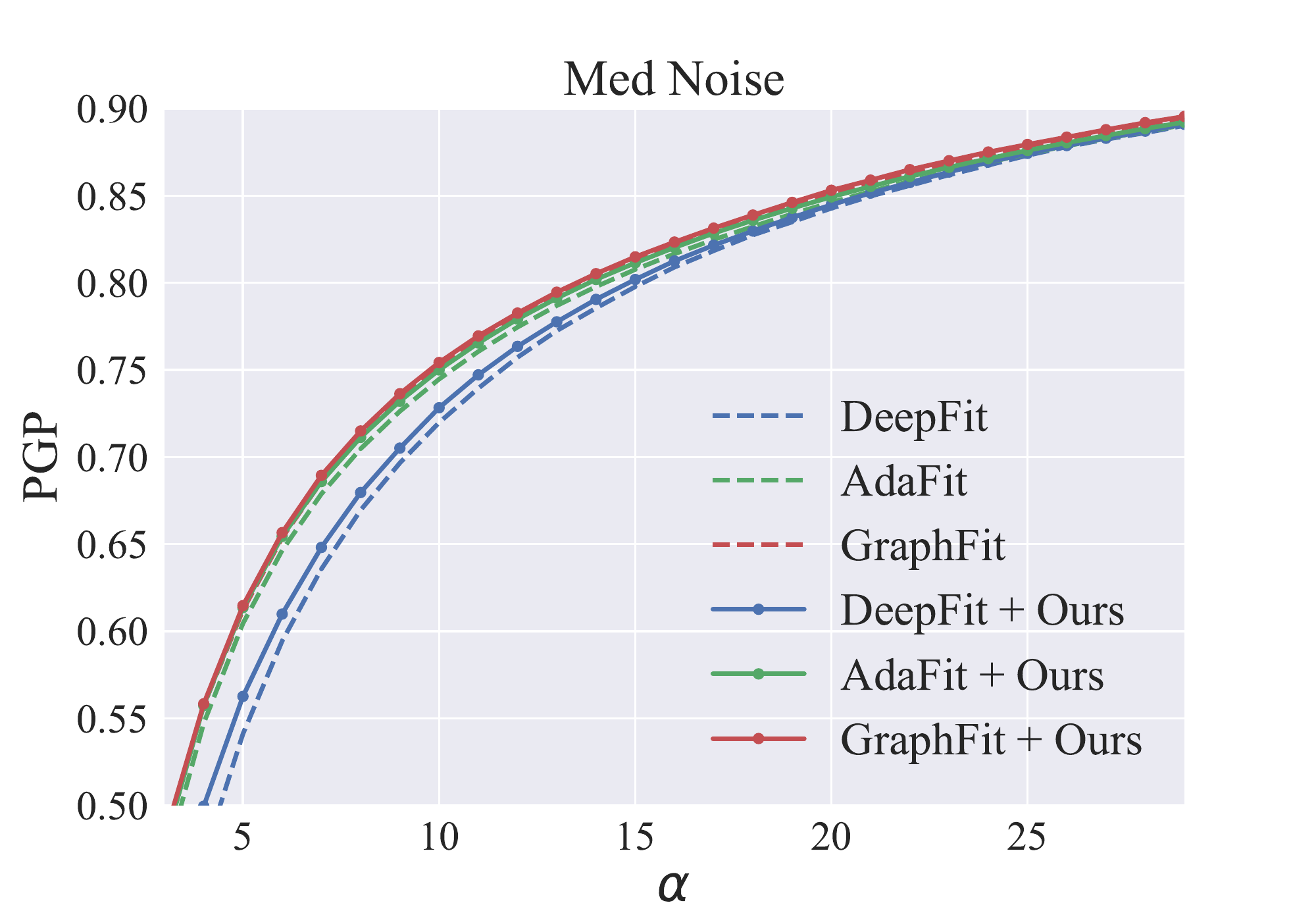}
\end{minipage} 
\begin{minipage}[t]{0.44\linewidth}
\includegraphics[height=3.cm]{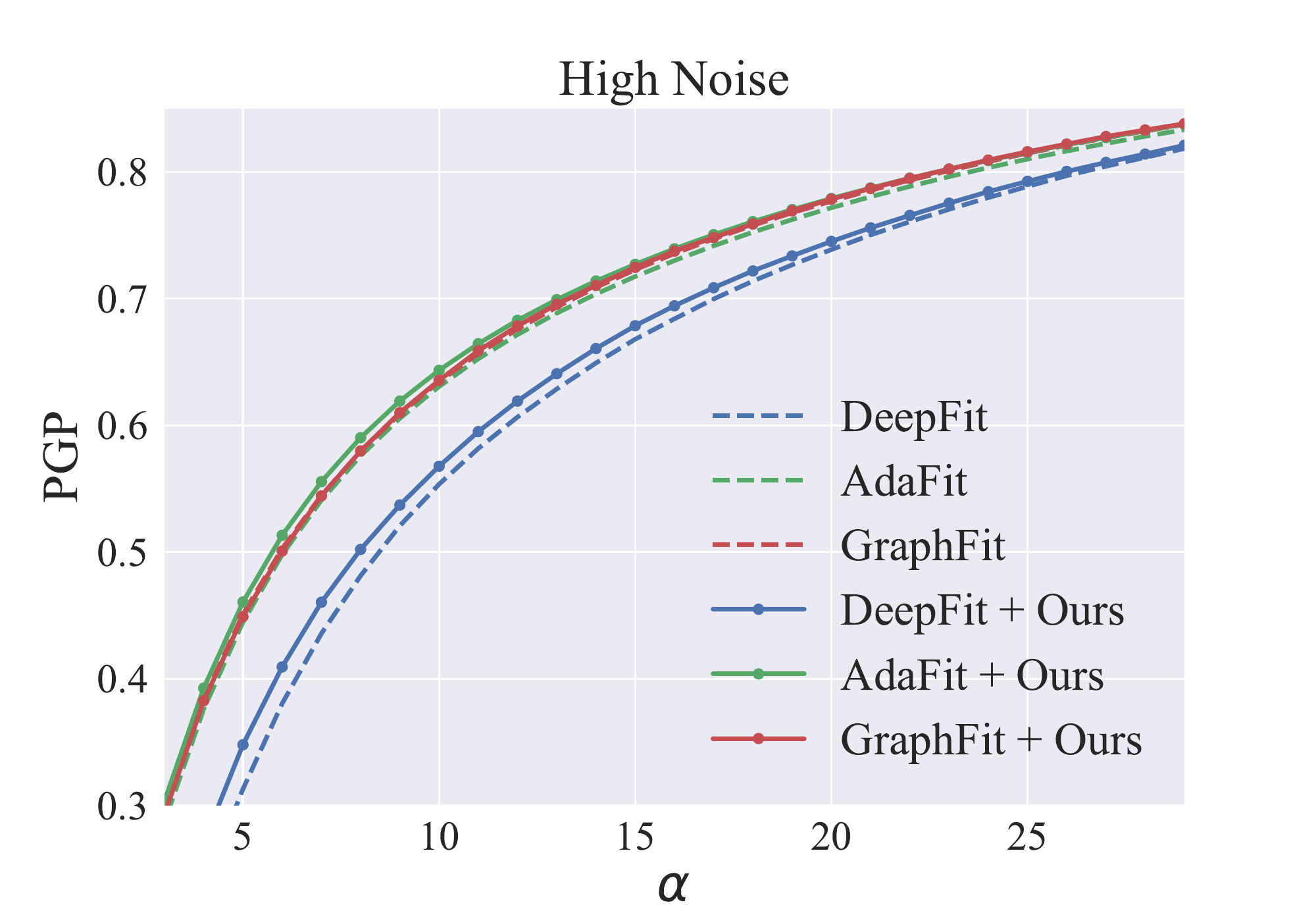}
\end{minipage}
\\
\begin{minipage}[t]{0.47\linewidth}
\includegraphics[height=3.cm]{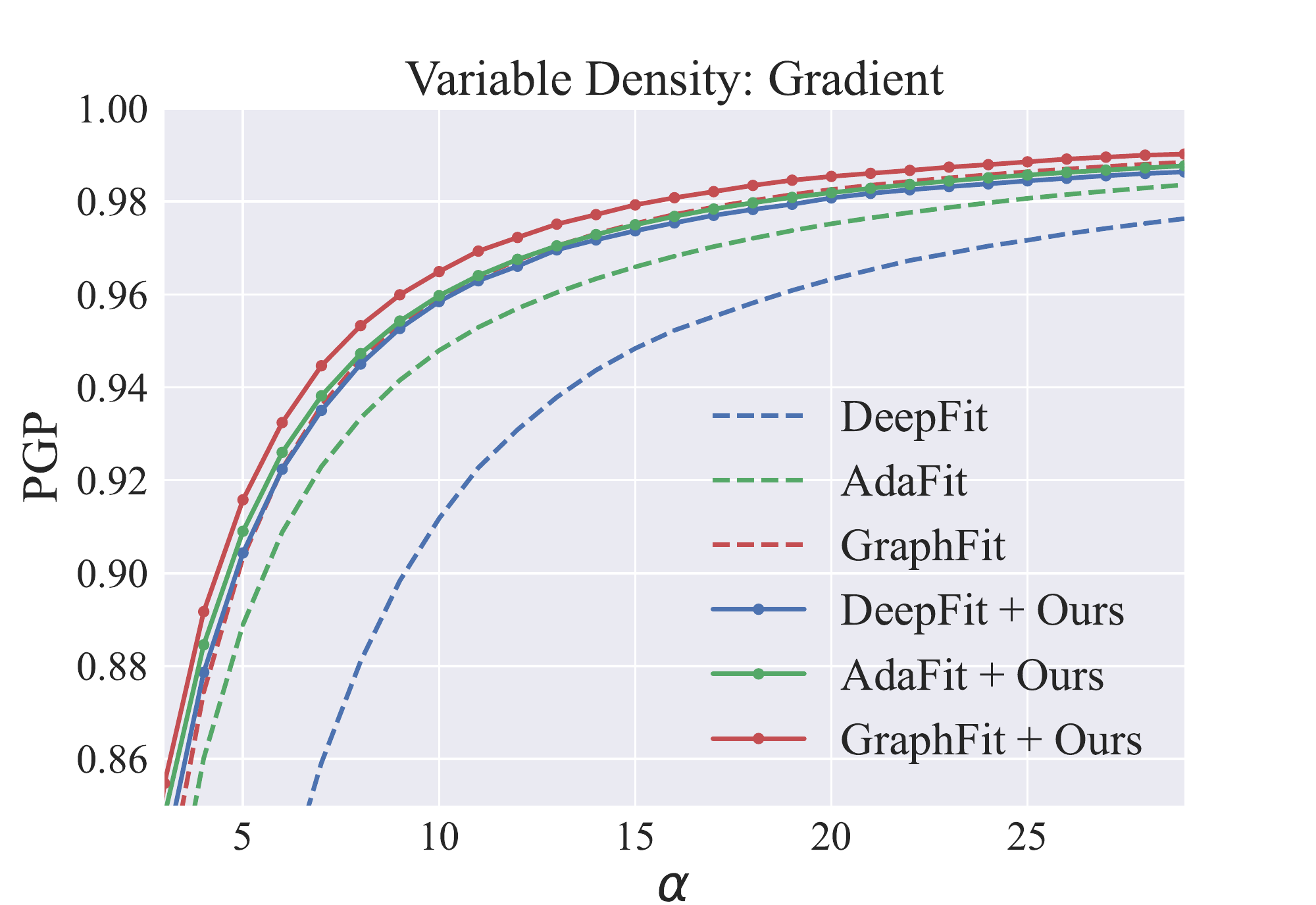}
\end{minipage} 
\begin{minipage}[t]{0.44\linewidth}
\includegraphics[height=3.cm]{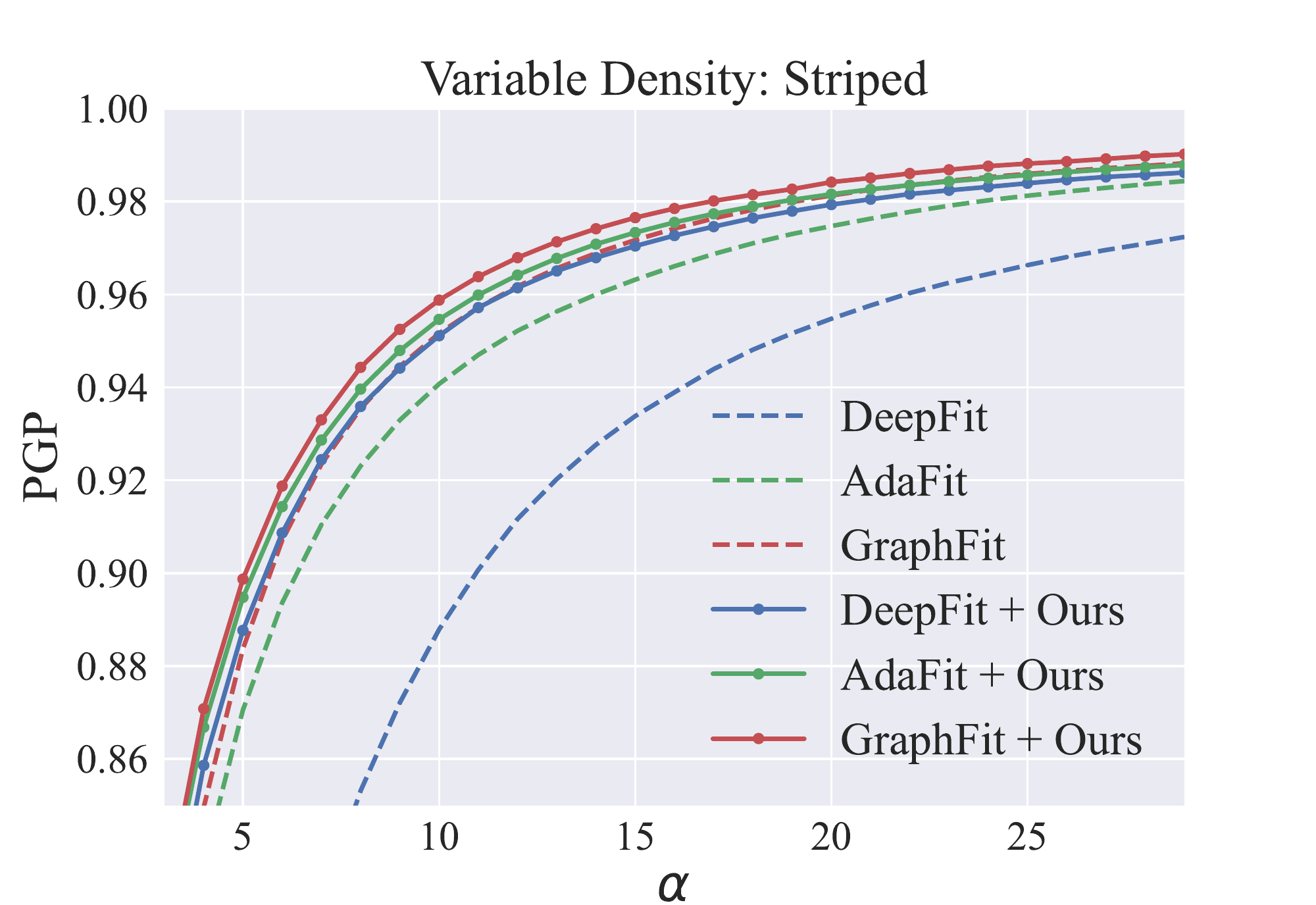}
\end{minipage}
\caption{Normal error AUC results of state-of-the-art models with or without our methods on PCPNet dataset.
The $x$ axis represents the threshold of error, and the $y$ axis represents the percentage of good points which have the lower error than a given threshold.}
\label{roc_pgp}
\vspace{-0.75em}
\end{figure}

\noindent\textbf{Quantitative results.} 
Table~\ref{sota_performance} reports the RMSE comparison of the exiting methods on PCPNet dataset.
The results imply that the proposed two basic ideas can be flexibly integrated with these polynomial surface fitting methods, and obtain evident improvements over the baseline counterparts. 
Moreover, as shown in Fig.~\ref{roc_pgp}, we further provide normal error Area Under the Curve (AUC) results of SOTA polynomial fitting models with or without our methods. 
Table~\ref{PGP_performance} also gives the quantitative results of PGP5 and PGP10 under no noise setting.  
From the results, we consistently improve the baseline models under different error thresholds, especially on the point clouds with density variations and low noise.  
The reason behind has two folds. 
First, the $z$-direction transformation enables to achieve a better surface fitting with a lower approximation error. 
Second, the normal error estimation can further update the rough estimated normal to a more accurate result.

\begin{figure}[t]
    \centering
    \includegraphics[height=5.4cm]{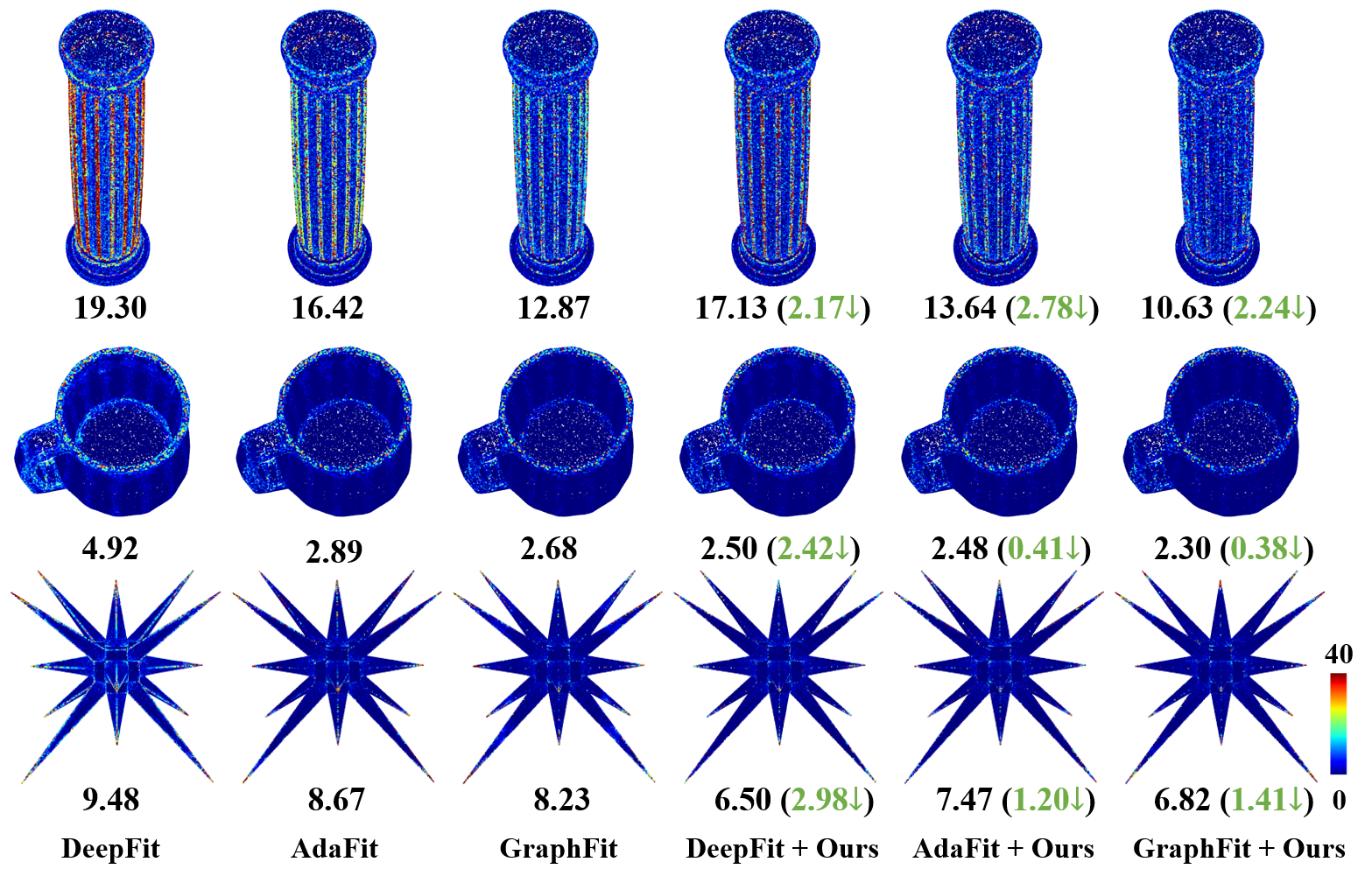}
    \caption{The error heatmap of normal estimation on PCPNet dataset. The bottom values represents the average error. The angle errors are mapped to a range from $0^\circ$ to $60^\circ$. }
    \label{fig_pcp}
    \vspace{-1em}
\end{figure}

\noindent\textbf{Qualitative results.} 
As shown in Fig.~\ref{fig_pcp},  we visualize the angle errors for the baseline models with or without the proposed methods.
From the results, we can observe our methods improve the robustness of baseline models on all the areas, such as curved regions and sharp edges. More visualization results can be found in the supplementary material.

\begin{table}[t]
    \begin{center}
    \caption{Comparison of percentage of good points PGP5 and PGP10 on  PCPNet dataset under no noise. Higher is better.  }
    \label{PGP_performance}
    \resizebox{0.9\linewidth}{!}{
    \begin{tabular}{c|cc|cc|cc}
    \toprule[1pt]
   {Method} & \multicolumn{2}{c|}{DeepFit} &\multicolumn{2}{c|}{AdaFit} &\multicolumn{2}{c}{GraphFit}\\
  &PGP5 &PGP10&PGP5 &PGP10&PGP5 &PGP10\\ 
    \midrule[0.5pt] 
 {Baseline} & 80.03&90.72&88.24 &94.36& 89.73&95.66\\
 {+ Ours} &89.83 &95.59& 90.40 &95.82&\textbf{91.28}&\textbf{96.59}\\
    \bottomrule[1pt]
    \end{tabular}}
    \end{center}
    \vspace{-1em}
\end{table}

\subsection{Comparison on Real-world Data} 
\begin{figure}[t]
    \centering
    \includegraphics[height=7.65cm]{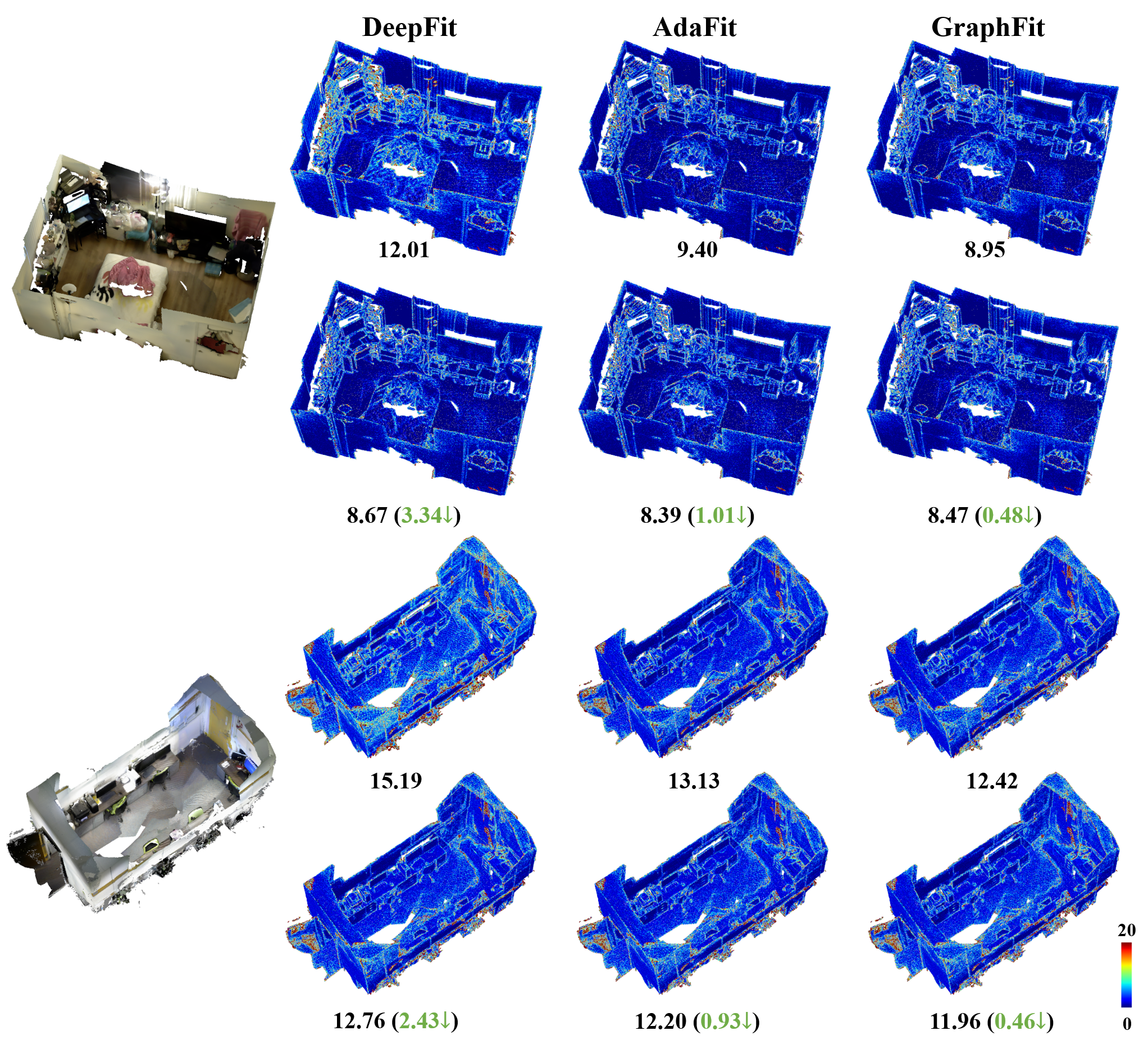}
    \caption{The error heatmap of normal estimation on sceneNN dataset. The bottom values denotes the average error. For each scene, the first line is produced by the baseline models, and the second line shows the results of applying our methods on the corresponding baseline. }
    \label{fig_scene}
    \vspace{-1em}
\end{figure}

In order to validate the proposed methods on real-world scenarios, we choose SceneNN~\cite{scenenn-3dv16} dataset for evaluation, which contains 76 scenes captured by a depth camera. 
We follow the settings of AdaFit~\cite{zhu2021adafit}  to obtain the sampled point clouds and ground-truth normals from provided ground-truth reconstructed meshes. 
The models trained on PCPNet dataset are directly utilized for evaluation. 
Table~\ref{SceneNN_comparsion} gives the quantitative results on all the scenes. The results show that our approaches also bring benefits to the baseline models on real-world data. 
Moreover, as shown in Fig.~\ref{fig_scene}, we randomly choose several scenes to visualize the normal errors. 
We can observe the real-world data is incomplete with many outliers and noise, which is a more challenging  than synthetic data.
Nevertheless, we can consistently improve the performance of SOTA models. 
Both quantitative and qualitative experiments demonstrate the good generalization ability of our methods on real-world data. 

\begin{table}[t]
    \begin{center}
    \caption{Normal angle RMSE on SceneNN dataset. }
    \label{SceneNN_comparsion}
    \resizebox{0.9\linewidth}{!}{
    \begin{tabular}{c|lll}
    \toprule[1pt]
  {Method} & DeepFit &AdaFit&GraphFit\\
    \midrule[0.5pt] 
 Baseline & 17.13& 15.49 & 14.79 \\
 + Ours & 14.57 (~\textcolor{green}{\textbf{2.56}$\downarrow$})& 14.45 (~\textcolor{green} {\textbf{1.04}$\downarrow$}) &  14.51 (~\textcolor{green}{\textbf{0.28}$\downarrow$})\\
    \bottomrule[1pt]
    \end{tabular}}
    \end{center}
    \vspace{-1em}
\end{table} 

\begin{table*}[t]
    \begin{center}
    \caption{ Normal angle RMSE of SOTA models with or without the proposed methods on PCPNet dataset. }
    \label{ablation_refine}
    \resizebox{0.9\linewidth}{!}{
    \begin{tabular}{c|cccc|cccc|cccc}
    \toprule[1pt]
  {Aug.} & \multicolumn{4}{c|}{DeepFit (size = 256)} &\multicolumn{4}{c|}{AdaFit (size = 700)}&\multicolumn{4}{c}{GraphFit (size = 500)}\\
    \midrule[0.5pt] 
 \textbf{$Z$-direction Trans.} &&\checkmark& &\checkmark& &\checkmark& &\checkmark& &\checkmark& &\checkmark\\
 \textbf{Error Estimation} &&&\checkmark&\checkmark& &&\checkmark &\checkmark& &&\checkmark &\checkmark\\
  No Noise &6.51 & 6.27 & 5.01 & 4.90 & 5.19 & 4.93 & 4.72 & 4.71&4.45&4.27&4.36&4.11\\
  Low Noise & 9.21 & 9.10 & 9.09 & 8.91 & 9.05 & 8.94 & 8.81 & 8.75&8.74&8.79&8.71&8.66\\
  Med Noise & 16.72 & 16.68 & 16.66 & 16.61 & 16.44 & 16.39 & 16.34 & 16.31&16.05&16.02&16.05&16.02\\
  High Noise & 23.12 & 22.98 & 22.94 & 22.87 & 21.94 & 21.56 & 21.80 & 21.64&21.64&21.66&21.60&21.57\\
  Gradient & 7.31 & 7.17 & 5.69 & 5.52 & 5.90 &  5.63 & 5.59 & 5.51&5.22&4.98 &5.06&4.83\\
  Striped & 7.92 & 7.73 & 5.85 & 5.70 & 6.01 & 5.89 &5.62 & 5.48&5.48&5.10&5.18&4.89\\
  \midrule[0.5pt]
  Average & 11.80 & 11.66 & 10.87 & \textbf{10.75} & 10.76 & 10.56 & 10.48 &\textbf{10.40}&10.26&10.14 &10.16&\textbf{10.01}\\
    \bottomrule[1pt]
    \end{tabular}}
    \end{center}
    \vspace{-1.5em}
\end{table*}

\begin{table}[t]
    \begin{center}
    \caption{Normal angle RMSE with different Jet order $n$ on PCPNet dataset. We choose single-scale AdaFit with 256 neighborhood size as baseline.  }
    \label{order_ablation}
    \resizebox{0.9\linewidth}{!}{
    \begin{tabular}{c|cc|cc|cc}
    \toprule[1pt]
  {Order} & \multicolumn{2}{c|}{1} &\multicolumn{2}{c|}{2} &\multicolumn{2}{c}{3}\\
    \midrule[0.5pt] 
 Baseline & \checkmark&& \checkmark&& \checkmark&\\
 + Ours & &\checkmark& &\checkmark& &\checkmark\\
  No Noise &5.13& 5.03 & 5.16 & 4.91 &  5.17  & 4.87 \\
  Low Noise  &9.23&  8.89&  9.11&  9.08& 9.17 &  9.02\\
  Med Noise  &16.76&  16.71 & 16.71 & 16.73 &16.71& 16.72 \\
  High Noise &22.92& 22.83 &  22.91 &  22.94&23.02&  22.87\\
  Gradient &5.87& 5.73 & 5.79  & 5.56  &6.03&  5.72\\
  Striped &6.01&  5.82&  5.92 & 5.71 &6.00& 5.79\\
  \midrule[0.5pt]
  Average &10.99 & 10.84 & 10.93 & 10.82&  11.02 & 10.83\\
    \bottomrule[1pt]
    \end{tabular}}
    \end{center}
    \vspace{-1em}
\end{table}

\begin{table}[t]
    \begin{center}
    \caption{Model complexity comparison. The inference time per point is tested on a NVIDIA TITAN X. }
    \label{complexity_performance}
    \resizebox{0.9\linewidth}{!}{
    \begin{tabular}{c|cccc}
    \toprule[1pt]
   {Method} &  \makecell[c]{Params \\ (M)} & \makecell[c]{Model size \\ (MB)} & \makecell[c]{Time \\ (ms)} & Avg. error\\
    \midrule[0.5pt] 
PCPNet~\cite{guerrero2018PCPNet} & 21.30 &85.41&-&14.56\\
Nesti-Net~\cite{ben2019nesti} & 170.10 &2,010.00 &-&12.41\\
 {DeepFit}~\cite{ben2020deepfit} & \textbf{3.44} &\textbf{13.53}&\textbf{0.47}&11.80\\
 {AdaFit}~\cite{zhu2021adafit} & 4.07&16.14&0.49&10.76\\
 {GraphFit}~\cite{li2022graphfit} & 4.16&16.38&2.07&10.26\\
  \midrule[0.5pt] 
  {DeepFit + Ours} & 3.64&14.26&0.56&10.75\\
 {AdaFit + Ours} & 4.28&16.85&0.58&10.40\\
 {GraphFit + Ours} & 4.36&17.09&2.19&\textbf{10.01}\\
    \bottomrule[1pt]
    \end{tabular}}
    \end{center}
    \vspace{-1em}
\end{table} 

\subsection{Ablation Study} 

To verify the effectiveness of our methods, we conduct extensive ablation studies on PCPNet dataset. 

\noindent\textbf{Influence of model components.} 
Firstly, we check the performance gain on DeepFit, AdaFit, and GraphFit, by integrating the proposed two methods with them. 
Table~\ref{ablation_refine} reports the ablation results. 
As we can see, all baseline models consistently obtain performance improvement after integrating with a solo version of our methods ($z$-direction transformation or normal error estimation). 
Moreover, they can achieve the best normal estimation results  with the combination of two model components. 
The results imply that our methods enable to reduce the approximation error of the fitting problem and improve the precision of surface normal. 
Note that we conduct above experiments under their optimal setting of input neighborhood size. 
So, the experimental results can also verify that our methods are robust to the neighborhood size, which can bring benefits to all baseline models under different settings. 

\noindent\textbf{Robustness against the Jet orders.} 
Moreover, we conduct experiments to verify the robustness of our methods against the polynomial order $n$. Considering the training consumption and computing resource, we choose the single-scale AdaFit as baseline, and the input neighborhood size is 256. The results in Table~\ref{order_ablation} show all the models have similar performance, indicating our methods can also work well under the different polynomial order $n$.

\subsection{Model Complexity Analysis} 
In addition, we make a comparison on model complexity. As given in Table~\ref{complexity_performance}, we only increase few extra burdens in terms of model complexity and bring evident improvement, indicating that the proposed methods are light-weight yet effective. 
For instance, after integrating with our methods, DeepFit is able to outperform the original AdaFit with fewer parameters. 
Thus, we can achieve a better practical implementation for the balance between the model accuracy and efficiency.

\section{Conclusion}
In this work, we study the approximation error in existing 3D surface fitting methods, and find there exists gaps between estimated and precise surface normals. To handle this problem, we propose two basic design principles, \ie, $z$-direction transformation and normal error estimation. The former is able to provide a better surface fitting with a lower approximation error, and the later can adjust the rough estimated normal to a more precise result. 
The proposed two principles can be flexibly integrated with the current SOTA polynomial surface fitting methods in a plug-and-play manner, and achieve significant improvements on both synthetic and real-world datasets.

{\small
\bibliographystyle{ieee_fullname}

\begin{thebibliography}{10}\itemsep=-1pt

\bibitem{alliez2007voronoi}
Pierre Alliez, David Cohen-Steiner, Yiying Tong, and Mathieu Desbrun.
\newblock Voronoi-based variational reconstruction of unoriented point sets.
\newblock In {\em Symposium on Geometry processing}, volume~7, pages 39--48,
  2007.

\bibitem{amenta1998surface}
Nina Amenta and Marshall Bern.
\newblock Surface reconstruction by voronoi filtering.
\newblock In {\em Proceedings of the fourteenth annual symposium on
  Computational geometry}, pages 39--48, 1998.

\bibitem{ben2020deepfit}
Yizhak Ben-Shabat and Stephen Gould.
\newblock Deepfit: 3d surface fitting via neural network weighted least
  squares.
\newblock In {\em European Conference on Computer Vision}, pages 20--34.
  Springer, 2020.

\bibitem{ben20183dmfv}
Yizhak Ben-Shabat, Michael Lindenbaum, and Anath Fischer.
\newblock 3dmfv: Three-dimensional point cloud classification in real-time
  using convolutional neural networks.
\newblock {\em IEEE Robotics and Automation Letters}, 3(4):3145--3152, 2018.

\bibitem{ben2019nesti}
Yizhak Ben-Shabat, Michael Lindenbaum, and Anath Fischer.
\newblock Nesti-net: Normal estimation for unstructured 3d point clouds using
  convolutional neural networks.
\newblock In {\em Proceedings of the IEEE/CVF Conference on Computer Vision and
  Pattern Recognition}, pages 10112--10120, 2019.

\bibitem{boulch2012fast}
Alexandre Boulch and Renaud Marlet.
\newblock Fast and robust normal estimation for point clouds with sharp
  features.
\newblock {\em Computer graphics forum}, 31(5):1765--1774, 2012.

\bibitem{boulch2016deep}
Alexandre Boulch and Renaud Marlet.
\newblock Deep learning for robust normal estimation in unstructured point
  clouds.
\newblock {\em Computer Graphics Forum}, 35(5):281--290, 2016.

\bibitem{cazals2005estimating}
Fr{\'e}d{\'e}ric Cazals and Marc Pouget.
\newblock Estimating differential quantities using polynomial fitting of
  osculating jets.
\newblock {\em Computer Aided Geometric Design}, 22(2):121--146, 2005.

\bibitem{dai2023MDR-MFI}
Weidong Dai, Xuejun Yan, Jingjing Wang, Di Xie, and Shiliang Pu.
\newblock Mdr-mfi:multi-branch decoupled regression and multi-scale feature
  interaction for partial-to-partial cloud registration.
\newblock In {\em IEEE International Conference on Acoustics, Speech and Signal
  Processing (ICASSP)}, 2023.

\bibitem{du2022point}
Hang Du, Xuejun Yan, Jingjing Wang, Di Xie, and Shiliang Pu.
\newblock Point cloud upsampling via cascaded refinement network.
\newblock In {\em Asian Conference on Computer Vision}, 2022.

\bibitem{edirimuni2022contrastive}
Dasith de~Silva Edirimuni, Xuequan Lu, Gang Li, and Antonio Robles-Kelly.
\newblock Contrastive learning for joint normal estimation and point cloud
  filtering.
\newblock {\em arXiv preprint arXiv:2208.06811}, 2022.

\bibitem{guennebaud2007algebraic}
Ga{\"e}l Guennebaud and Markus Gross.
\newblock Algebraic point set surfaces.
\newblock In {\em ACM siggraph 2007 papers}, pages 23--es, 2007.

\bibitem{guerrero2018PCPNet}
Paul Guerrero, Yanir Kleiman, Maks Ovsjanikov, and Niloy~J Mitra.
\newblock Pcpnet learning local shape properties from raw point clouds.
\newblock {\em Computer Graphics Forum}, 37(2):75--85, 2018.

\bibitem{hashimoto2019normal}
Taisuke Hashimoto and Masaki Saito.
\newblock Normal estimation for accurate 3d mesh reconstruction with point
  cloud model incorporating spatial structure.
\newblock In {\em Proceedings of the IEEE/CVF Conference on Computer Vision and
  Pattern Recognition workshops}, volume~1, 2019.

\bibitem{hoppe1992surface}
Hugues Hoppe, Tony DeRose, Tom Duchamp, John McDonald, and Werner Stuetzle.
\newblock Surface reconstruction from unorganized points.
\newblock In {\em Proceedings of the 19th Annual conference on Computer
  Graphics and Interactive Techniques}, pages 71--78, 1992.

\bibitem{scenenn-3dv16}
Binh-Son Hua, Quang-Hieu Pham, Duc~Thanh Nguyen, Minh-Khoi Tran, Lap-Fai Yu,
  and Sai-Kit Yeung.
\newblock Scenenn: A scene meshes dataset with annotations.
\newblock In {\em International Conference on 3D Vision (3DV)}, 2016.

\bibitem{huang2021predator}
Shengyu Huang, Zan Gojcic, Mikhail Usvyatsov, Andreas Wieser, and Konrad
  Schindler.
\newblock Predator: Registration of 3d point clouds with low overlap.
\newblock In {\em Proceedings of the IEEE/CVF Conference on computer vision and
  pattern recognition}, pages 4267--4276, 2021.

\bibitem{kazhdan2006poisson}
Michael Kazhdan, Matthew Bolitho, and Hugues Hoppe.
\newblock Poisson surface reconstruction.
\newblock In {\em Proceedings of the fourth Eurographics symposium on Geometry
  processing}, volume~7, 2006.

\bibitem{lenssen2020deep}
Jan~Eric Lenssen, Christian Osendorfer, and Jonathan Masci.
\newblock Deep iterative surface normal estimation.
\newblock In {\em Proceedings of the IEEE/CVF Conference on Computer Vision and
  Pattern Recognition}, pages 11247--11256, 2020.

\bibitem{levin1998approximation}
David Levin.
\newblock The approximation power of moving least-squares.
\newblock {\em Mathematics of computation}, 67(224):1517--1531, 1998.

\bibitem{li2022graphfit}
Keqiang Li, Mingyang Zhao, Huaiyu Wu, Dong-Ming Yan, Zhen Shen, Fei-Yue Wang,
  and Gang Xiong.
\newblock Graphfit: Learning multi-scale graph-convolutional representation for
  point cloud normal estimation.
\newblock In {\em European Conference on Computer Vision}, pages 651--667.
  Springer, 2022.

\bibitem{li2022hsurf}
Qing Li, Yu-Shen Liu, Jin-San Cheng, Cheng Wang, Yi Fang, and Zhizhong Han.
\newblock Hsurf-net: Normal estimation for 3d point clouds by learning hyper
  surfaces.
\newblock In {\em Advances in Neural Information Processing Systems}, 2022.

\bibitem{lu2020deep}
Dening Lu, Xuequan Lu, Yangxing Sun, and Jun Wang.
\newblock Deep feature-preserving normal estimation for point cloud filtering.
\newblock {\em Computer-Aided Design}, 125:102860, 2020.

\bibitem{lu2020low}
Xuequan Lu, Scott Schaefer, Jun Luo, Lizhuang Ma, and Ying He.
\newblock Low rank matrix approximation for 3d geometry filtering.
\newblock {\em IEEE Transactions on Visualization and Computer Graphics}, 2020.

\bibitem{ma2021neural}
Baorui Ma, Zhizhong Han, Yu-Shen Liu, and Matthias Zwicker.
\newblock Neural-pull: Learning signed distance functions from point clouds by
  learning to pull space onto surfaces.
\newblock In {\em Proceedings of the 38th International Conference on Machine
  Learning}, volume 139, 2021.

\bibitem{merigot2010voronoi}
Quentin M{\'e}rigot, Maks Ovsjanikov, and Leonidas~J Guibas.
\newblock Voronoi-based curvature and feature estimation from point clouds.
\newblock {\em IEEE Transactions on Visualization and Computer Graphics},
  17(6):743--756, 2010.

\bibitem{pomerleau2015review}
Fran{\c{c}}ois Pomerleau, Francis Colas, Roland Siegwart, et~al.
\newblock A review of point cloud registration algorithms for mobile robotics.
\newblock {\em Foundations and Trends{\textregistered} in Robotics},
  4(1):1--104, 2015.

\bibitem{qi2017pointnet}
Charles~R Qi, Hao Su, Kaichun Mo, and Leonidas~J Guibas.
\newblock Pointnet: Deep learning on point sets for 3d classification and
  segmentation.
\newblock In {\em Proceedings of the IEEE/CVF Conference on Computer Vision and
  Pattern Recognition}, pages 652--660, 2017.

\bibitem{qi2017pointnet++}
Charles~Ruizhongtai Qi, Li Yi, Hao Su, and Leonidas~J Guibas.
\newblock Pointnet++: Deep hierarchical feature learning on point sets in a
  metric space.
\newblock In {\em Advances in Neural Information Processing Systems},
  volume~30, 2017.

\bibitem{qian2022pointnext}
Guocheng Qian, Yuchen Li, Houwen Peng, Jinjie Mai, Hasan Abed Al~Kader Hammoud,
  Mohamed Elhoseiny, and Bernard Ghanem.
\newblock Pointnext: Revisiting pointnet++ with improved training and scaling
  strategies.
\newblock In {\em Advances in Neural Information Processing Systems}, 2022.

\bibitem{qian2020pugeo}
Yue Qian, Junhui Hou, Sam Kwong, and Ying He.
\newblock Pugeo-net: A geometry-centric network for 3d point cloud upsampling.
\newblock In {\em European Conference on Computer Vision}, pages 752--769.
  Springer, 2020.

\bibitem{wang2022deep}
Shiyao Wang, Xiuping Liu, Jian Liu, Shuhua Li, and Junjie Cao.
\newblock Deep patch-based global normal orientation.
\newblock {\em Computer-Aided Design}, page 103281, 2022.

\bibitem{wang2019deep}
Yue Wang and Justin~M Solomon.
\newblock Deep closest point: Learning representations for point cloud
  registration.
\newblock In {\em Proceedings of the IEEE/CVF international conference on
  computer vision}, pages 3523--3532, 2019.

\bibitem{wang2019dynamic}
Yue Wang, Yongbin Sun, Ziwei Liu, Sanjay~E Sarma, Michael~M Bronstein, and
  Justin~M Solomon.
\newblock Dynamic graph cnn for learning on point clouds.
\newblock {\em Acm Transactions On Graphics (tog)}, 38(5):1--12, 2019.

\bibitem{yan2022fbnet}
Xuejun Yan, Hongyu Yan, Jingjing Wang, Hang Du, Zhihong Wu, Di Xie, Shiliang
  Pu, and Li Lu.
\newblock Fbnet: Feedback network for point cloud completion.
\newblock In {\em European Conference on Computer Vision}, pages 676--693.
  Springer, 2022.

\bibitem{zhang2022geometry}
Jie Zhang, Jun-Jie Cao, Hai-Rui Zhu, Dong-Ming Yan, and Xiu-Ping Liu.
\newblock Geometry guided deep surface normal estimation.
\newblock {\em Computer-Aided Design}, 142:103119, 2022.

\bibitem{zhou2020geometry}
Haoran Zhou, Honghua Chen, Yidan Feng, Qiong Wang, Jing Qin, Haoran Xie, Fu~Lee
  Wang, Mingqiang Wei, and Jun Wang.
\newblock Geometry and learning co-supported normal estimation for unstructured
  point cloud.
\newblock In {\em Proceedings of the IEEE/CVF Conference on Computer Vision and
  Pattern Recognition}, pages 13238--13247, 2020.

\bibitem{zhou2022refine}
Haoran Zhou, Honghua Chen, Yingkui Zhang, Mingqiang Wei, Haoran Xie, Jun Wang,
  Tong Lu, Jing Qin, and Xiao-Ping Zhang.
\newblock Refine-net: Normal refinement neural network for noisy point clouds.
\newblock {\em IEEE Transactions on Pattern Analysis and Machine Intelligence},
  2022.

\bibitem{zhou2020normal}
Jun Zhou, Hua Huang, Bin Liu, and Xiuping Liu.
\newblock Normal estimation for 3d point clouds via local plane constraint and
  multi-scale selection.
\newblock {\em Computer-Aided Design}, 129:102916, 2020.

\bibitem{zhou2022fast}
Jun Zhou, Wei Jin, Mingjie Wang, Xiuping Liu, Zhiyang Li, and Zhaobin Liu.
\newblock Fast and accurate normal estimation for point clouds via patch
  stitching.
\newblock {\em Computer-Aided Design}, 142:103121, 2022.

\bibitem{zhu2022semi}
Runsong Zhu, Di Kang, Ka-Hei Hui, Yue Qian, Xuefei Zhe, Zhen Dong, Linchao Bao,
  and Chi-Wing Fu.
\newblock Semi-signed neural fitting for surface reconstruction from unoriented
  point clouds.
\newblock {\em arXiv preprint arXiv:2206.06715}, 2022.

\bibitem{zhu2021adafit}
Runsong Zhu, Yuan Liu, Zhen Dong, Yuan Wang, Tengping Jiang, Wenping Wang, and
  Bisheng Yang.
\newblock Adafit: Rethinking learning-based normal estimation on point clouds.
\newblock In {\em Proceedings of the IEEE/CVF International Conference on
  Computer Vision}, pages 6118--6127, 2021.

\end{thebibliography}

}

\newpage

\appendix

\section{Supplementary Materials}
In the following, we provide implementation details of network architecture. 
For a thorough evaluation, we conduct more experiments, including ablation studies on the $z$-direction transformation loss, robustness against neighborhood sizes and Jet orders $n$, iterative estimation experiment and more visualization results. In addition, we also provide an application of normal estimation, \ie surface reconstruction, to further verify the effectiveness of our methods.

\section{Network Architecture Details}
In this section, we provide the details of network architecture.  

\subsection{GCN-based Transformation Network} 
\begin{figure}[ht]
    \centering
    \includegraphics[height=2.5cm]{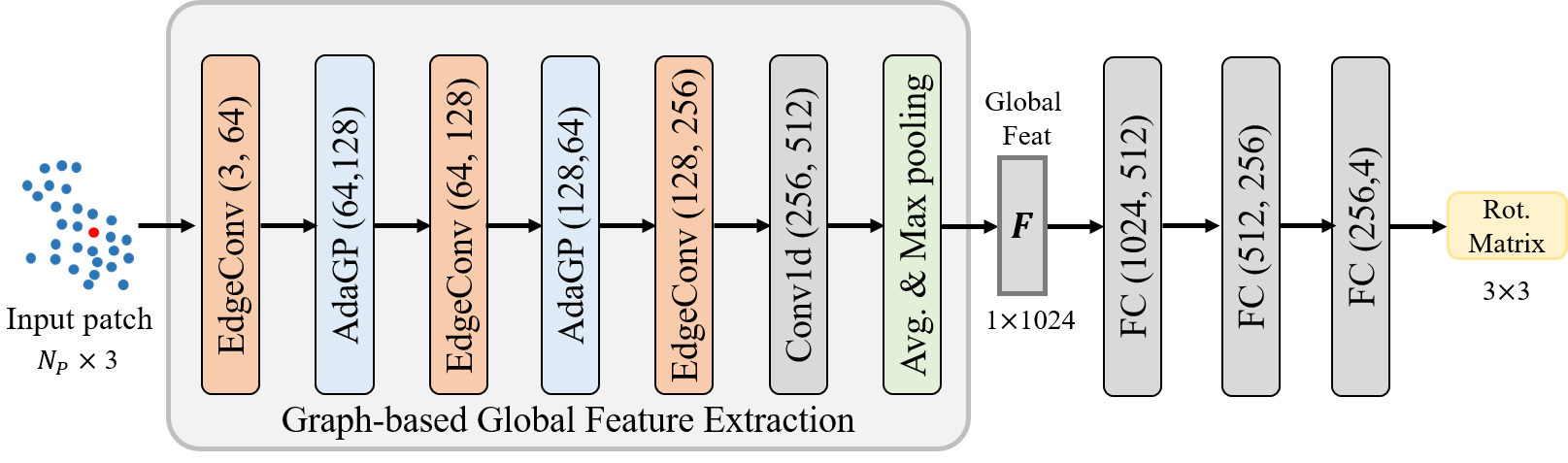}
    \caption{Detailed architecture of GCN-based spatial transformation network. }
    \label{gcn_model}
\end{figure}

In order to accomplish $z$-direction transformation, we design a GCN-based spatial transformation network, which is shown in Fig.~\ref{gcn_model}. 
In particular, EdgeConv (3, 64) denotes a EdgeConv~\cite{wang2019dynamic} layer with the number of input channel as 3, and the number of output channel as 64. AdaGP (64, 128) represents an adaptive graph pooling~\cite{yan2022fbnet} layer with the number of input/output channel as 64, and the output number of point as 128.
Conv1d (256, 512) indicates a 1D convolutional layer with the number of input channel as 256, and the number of output channel as 512. 
Avg. and Max pooling denotes a combination of average and max pooling operations. 
FC (1024, 512) indicates a fully-connected layer with the number of input channel as 1024, and the number of output channel as 512.

\subsection{Normal Error Estimation}  
\begin{figure}[ht]
    \centering
    \includegraphics[height=3.5cm]{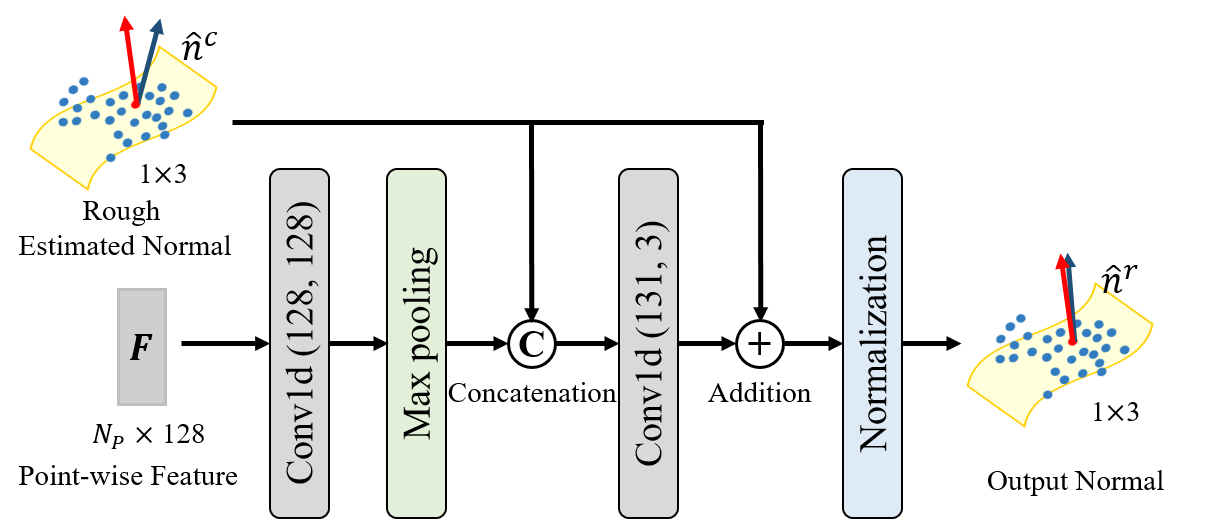}
    \caption{Architecture details of the normal error estimation network. }
    \label{error_model}
\end{figure}

The network architecture of the normal error estimation is shown in Fig.~\ref{error_model}. Specifically, the point-wise feature is fed into a 1D convolutional layer with the number of input/output channel as 128. Subsequently, through a max pooling layer, the global feature is concatenated with the rough estimated normal to estimate the normal error.
Then, the error of normal estimation is added on the rough estimation to obtain the output normal. 
Finally, the output normal is normalized into a unit vector.

\begin{table*}[t]
    \begin{center}
    \caption{Normal angle RMSE with different neighborhood size on PCPNet dataset.}
    \label{size_ablation}
    \resizebox{1\linewidth}{!}{
    \begin{tabular}{c|cc|cc|cc|cc|cc|cc|cc|cc|cc}
    \toprule[1pt]
     & \multicolumn{6}{c|}{DeepFit}  & \multicolumn{6}{c|}{AdaFit} & \multicolumn{6}{c}{GraphFit} \\
  {Size} & \multicolumn{2}{c}{256} &\multicolumn{2}{c}{500} &\multicolumn{2}{c|}{700} &\multicolumn{2}{c}{256} &\multicolumn{2}{c}{500} &\multicolumn{2}{c|}{700} &  \multicolumn{2}{c}{256} &\multicolumn{2}{c}{500} &\multicolumn{2}{c}{700}\\
    \midrule[0.5pt] 
 Baseline & \checkmark& & \checkmark& & \checkmark&  & \checkmark& & \checkmark& & \checkmark&  & \checkmark& & \checkmark& & \checkmark& \\
 + Ours & &\checkmark& &\checkmark& &\checkmark& &\checkmark& &\checkmark& &\checkmark& &\checkmark& &\checkmark& &\checkmark\\
  None &6.51&4.90&7.10&5.67&7.35& 5.85&5.17&4.87&5.79& 5.33&5.19&4.71 &4.49&4.10&4.45&4.11&4.83&4.34\\
  Low  &9.21&8.91&9.41&9.23&9.63& 9.27&9.17&9.02&9.17& 9.16 &9.05&8.75 &8.80&8.78&8.74&8.66&8.70&8.68\\
  Med  &16.72&16.61&16.46&16.38&16.39&16.34&16.71&16.72&16.47& 16.43 &16.44& 16.31 &16.54&16.46&16.05&16.02&16.04&16.07\\
  High &23.12&22.87&21.97&21.82&21.74 &21.70 &23.02&22.87&22.12&21.91 &21.94& 21.64 &22.69&22.64&21.64&21.57&21.36&21.45\\
  Gradient &7.31&5.52&7.71&6.22&8.06& 6.43&6.03&5.72&6.64&5.84 &5.90& 5.51 &5.15&4.91&5.22&4.83&5.51&5.31\\
  Striped &7.92&5.70&8.66&6.51&9.26& 6.92&6.00&5.79&6.30&6.01&6.01& 5.48 &5.28&5.00&5.48&4.89&5.61&5.35\\
  \midrule[0.5pt]
  Average & 11.80 & 10.75  & 11.89 & 10.97& 12.07 &11.09  &11.02& 10.83& 11.08 & 10.78 &  10.76 & 10.40 &10.49 &10.33&10.26&10.01&10.34& 10.20\\
    \bottomrule[1pt]
    \end{tabular}}
    \end{center}
\end{table*} 

\begin{table*}[t]
    \begin{center}
    \caption{Normal angle RMSE with different Jet order $n$ on PCPNet dataset.}
    \label{order_ablation_v2}
    \resizebox{0.85\linewidth}{!}{
    \begin{tabular}{c|cc|cc|cc|cc|cc|cc}
    \toprule[1pt]
     & \multicolumn{6}{c|}{DeepFit} & \multicolumn{6}{c}{GraphFit} \\
  {Order} & \multicolumn{2}{c}{1} &\multicolumn{2}{c}{2} &\multicolumn{2}{c|}{3} & \multicolumn{2}{c}{1} &\multicolumn{2}{c}{2} &\multicolumn{2}{c}{3}\\
    \midrule[0.5pt] 
 Baseline & \checkmark& & \checkmark& & \checkmark&  & \checkmark& & \checkmark& & \checkmark& \\
 + Ours & &\checkmark& &\checkmark& &\checkmark& &\checkmark& &\checkmark& &\checkmark\\
  No Noise & 6.72& 5.25 & 8.08& 5.20&  6.51  & 4.90 &4.70& 4.22 & 4.62 & 4.17 &  4.45  & 4.11 \\
  Low Noise  & 9.55& 9.31  & 9.74 &  9.01& 9.21 &  8.91 &8.79& 8.75 & 8.71 & 8.79 &  8.74  & 8.66\\
  Med Noise  &16.77 &   16.77 & 16.56 & 16.61 &16.72& 16.61  &16.29& 16.30 & 16.11 & 16.02 & 16.05  & 16.02\\
  High Noise & 23.16 &  23.00 &  23.00 &  22.78&23.12&  22.87  &21.75& 21.72 & 21.78 & 21.66 &  21.64  & 21.57\\
  Gradient & 7.46&5.97  & 8.77  & 5.85  &7.31&  5.52  &5.26& 5.11 & 5.39 & 4.98 & 5.22  &4.83\\
  Striped &7.84 & 6.15 &  9.21 & 6.14 &7.92& 5.70  &5.44& 5.25 & 5.50 & 5.10 & 5.48 & 4.89\\
  \midrule[0.5pt]
  Average &11.92  & 11.08  & 12.56 & 10.93 &  11.80 & \textbf{10.75}  &10.37 & 10.24  & 10.34 & 10.12 &  10.26 & \textbf{10.01}\\
    \bottomrule[1pt]
    \end{tabular}}
    \end{center}
\end{table*}

\begin{table*}[t]
    \begin{center}
    \caption{Normal angle RMSE with different weights of $z$-direction transformation loss on PCPNet dataset. The left shows the results of directly applying the $z$-direction transformation loss on original quaternion spatial transformation network (Q-STN)~\cite{ben2020deepfit}, and the right is produced by using our GCN-based transformation network and proposed loss function.}
    \label{weight_ablation}
    \resizebox{0.85\linewidth}{!}{
    \begin{tabular}{c|cccccc|cccccc}
    \toprule[1pt]
     & \multicolumn{6}{c|}{DeepFit}& \multicolumn{6}{c}{DeepFit + Ours}\\
  {Trans. Weight} & 0&0.1& 1.0 &2.0&3.0&10.0& 0&0.1& 1.0 &2.0&3.0&10.0\\
    \midrule[0.5pt] 
  No Noise &6.51&6.49& 6.58&6.55&6.69& 6.76 &5.05&5.07&4.97&4.90&4.99&5.05\\
  Low Noise   &9.21&9.16&9.12& 9.10& 9.07&9.09 &9.09&9.05&8.94&8.91 &8.98&9.03\\
  Med Noise  &16.72&16.60&16.65& 16.64& 16.65 & 16.63 &16.74&16.67&16.66&16.61 &16.68&16.70\\ 
  High Noise  &23.12&23.02&22.96& 23.03& 23.02  &23.05 &22.88&22.87&22.86&22.87 &22.90&22.90\\
  Gradient  &7.31&7.29&7.46& 7.47& 7.51 &7.48 &5.79&5.82&5.69&5.52 &5.74&5.75\\
  Striped  &7.92&7.85&7.98&7.90& 7.95&7.96 &5.95&5.99&5.87&5.70&5.95&6.05 \\
  \midrule[0.5pt]
  Average  &11.80&\textbf{11.74}&11.79&11.78&11.82 &11.83&10.92&10.91&10.83&\textbf{10.75}  &10.87&10.91\\
    \bottomrule[1pt]
    \end{tabular}}
    \end{center}
\end{table*}

\section{More Experimental Results} 
In this section, we first conduct more ablation studies on the proposed methods. 
Then, we provide an application of normal estimation to surface reconstruction.
Finally, we give more visualization results.  

\subsection{Ablation Study}

\noindent\textbf{Robustness to the neighborhood size.} In the main text, we have reported the results of DeepFit~\cite{ben2020deepfit}, AdaFit~\cite{zhu2021adafit}, and GraphFit~\cite{li2022graphfit} under their optimal input neighborhood sizes. To further verify the robustness to the neighborhood sizes, we conduct comprehensive experiments on these baseline methods. As shown in Table~\ref{size_ablation}, we can find that, whatever the input size and the baseline model, an evident improvement can be obtained by our methods. 
The results demonstrate that the proposed methods are robust and effective under different neighborhood sizes.

\noindent\textbf{Robustness against the Jet orders.} 
In addition, we provide more results on DeepFit~\cite{ben2020deepfit} and GraphFit~\cite{li2022graphfit} to verify the robustness of our methods against the polynomial order $n$. Note that we have given the results of AdaFit ~\cite{zhu2021adafit} in the main text.
Here, as shown in Table.~\ref{order_ablation_v2}, we can obtain stable performance improvements over the baseline models under different polynomial orders. 
Besides, we observe that $n=3$ consistently achieves the best performance in terms of average normal angle RMSE. We consider the order $3$ is suitable for most points, and our methods can reduce the approximation error of normal estimation, which bring benefits to all the baseline models under different polynomial orders.

\noindent\textbf{Ablation on $Z$-direction Transformation Loss.} 
As presented in Sec.~\textcolor{red}{4.2} of the main text, we propose a $z$-direction transformation loss that constrains the transformation matrix to narrow the angle between the rotated ground-truth normal and the axis $z$. 
Here, we conduct an ablation study on $z$-direction transformation loss, to validate the effectiveness of our GCN-based transformation network and proposed loss function.  
First, we directly apply the $z$-direction transformation loss on original DeepFit model~\cite{ben2020deepfit}. 
From the results in Table~\ref{weight_ablation}, we can find that the expected transformation is non-trivial for the previous transformation network, and thus there is no obvious improvements on original DeepFit model~\cite{ben2020deepfit} (the left half of the table). 
Then, we replace the previous transformation network with our proposed GCN-based transformation network. 
In such a scenario, we can achieve evident performance improvements compared with baseline counterpart, and the best performance is obtained when the loss weight is set as 2. 
The experimental results imply that our $z$-direction transformation loss works well within a certain range.

\begin{table}[!ht]
    \begin{center}
    \caption{Iterative estimation on DeepFit model. The inference time is tested on a NVIDIA TITAN X. }
    \label{Iterative}
    \resizebox{0.85\linewidth}{!}{
    \begin{tabular}{c|c|c}
    \toprule[1pt]
  {Method}& Average RMSE &Time (ms)\\
    \midrule[0.5pt] 
  DeepFit &11.80 &\textbf{0.47}\\
  DeepFit (iterative) & 11.72&0.73 \\
  DeepFit + Ours &\underline{10.75} &\underline{0.56}\\
  DeepFit + Ours (iterative) &\textbf{10.72} &0.84\\ 
    \bottomrule[1pt]
    \end{tabular}}
    \end{center}
\end{table} 

\subsection{Iterative Estimation}
In the main text, we argue a better $z$-alignment could improve the precision of normal estimation.
Here, we conduct an iterative experiment to verify it.
The iterative estimation refers to feeding the normal results of $n$-jet fitting or error estimation module back to $n$-jet fitting for an estimation again. 
By doing so, we can rotate the estimated normal to the axis $z$ for a better $z$ alignment, and thus achieve a more accurate surface fitting. 
In Table~\ref{Iterative}, the results show a better $z$-alignment indeed reduces the error of normal estimation. 
However, simply iterative DeepFit (the 2nd line) still performs worse than our methods (the 3rd line). Besides, such scheme will cost much more inference time.

\subsection{Surface Reconstruction Application}
\begin{figure*}[t]
    \centering
    \includegraphics[height=19cm]{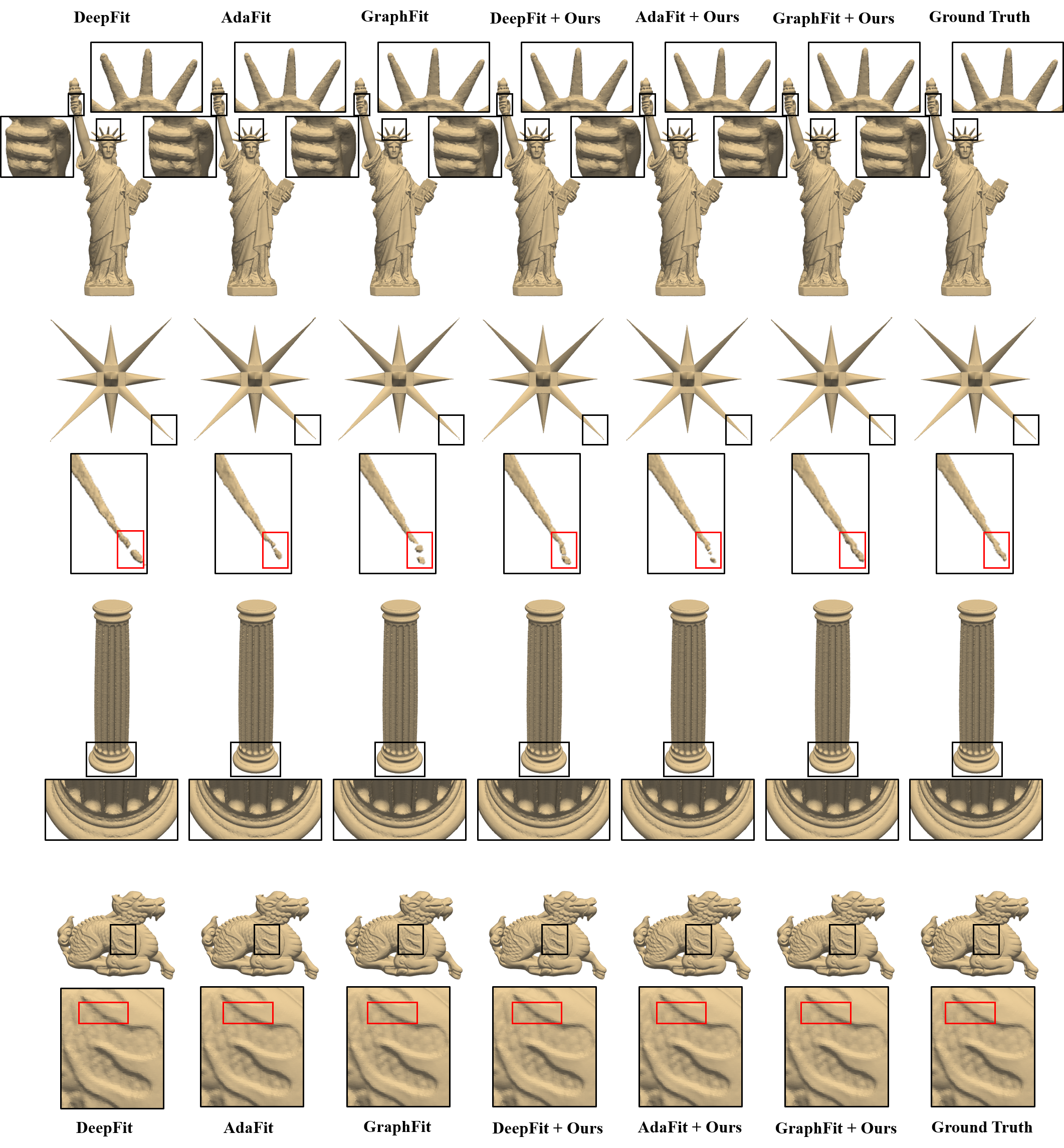}
    \caption{Comparison of surface reconstruction using the normals estimated by SOTA models with our methods. Our methods improve the baseline models to produce more accurate and complete surfaces.  }
    \label{fig_surface}
\end{figure*}

Accurate surface normals can benefit to reconstruct a better surface. So, we adopt the Poisson reconstruction implemented by Open3D library to  reconstruct the surface from the point cloud with the estimated normals. Fig.~\ref{fig_surface} shows that  accurate normals are helpful to reconstruct a more high-quality and complete surface from point clouds, such as the finger of liberty, and the sharp corner of star. Besides, we follow a common way~\cite{ma2021neural} that samples $1\times10^5$ points from the reconstructed meshes and computes the L2-CD distance of them.  
The quantitative results are given in Table~\ref{surface_reconstruction}. 
The results show that our methods can help the baseline models to obtain a better reconstructed surface in most cases, and consistently achieve improvements in terms of average reconstruction error. 

\begin{table}[t]
    \begin{center}
    \caption{L2-CD ($\times10^4$) comparison for surface reconstruction of baseline models with or without our methods.}
    \label{surface_reconstruction}
    \resizebox{1\linewidth}{!}{
    \begin{tabular}{c|ccc|ccc}
    \toprule[1pt]
     & {DeepFit}& {AdaFit}& {GraphFit}& \makecell[c]{DeepFit \\+ Ours}& \makecell[c]{AdaFit\\+ Ours}& \makecell[c]{GraphFit \\+ Ours}\\
    \midrule[0.5pt] 
  Liberty&0.290&0.180&0.142&0.117&0.096& 0.091\\
  Star sharp&0.148&0.106&0.105&0.099&0.112&0.101 \\
  Column&0.212&0.195&0.212&0.218&0.193& 0.179\\
  Netsuke&0.295&0.248&0.239&0.261&0.245&0.258\\
  \midrule[0.5pt]
  Average  &0.236&0.182&0.175&\makecell[c]{0.174\\ (~\textcolor{green}{\textbf{0.062}$\downarrow$})}&\makecell[c]{0.161\\ (~\textcolor{green}{\textbf{0.021}$\downarrow$})}&\makecell[c]{0.157 \\(~\textcolor{green}{\textbf{0.018}$\downarrow$})}\\
    \bottomrule[1pt]
    \end{tabular}}
    \end{center}
\end{table}

\subsection{Visualization Results}
In addition, we present more visualization results on PCPNet~\cite{guerrero2018PCPNet} dataset. Besides, we also provide the corresponding normal RMSE, and  the percentage of good points (PGP10 and PGP5) of each shape. 

\begin{figure*}[t!]
    \centering
    \includegraphics[height=18.25cm]{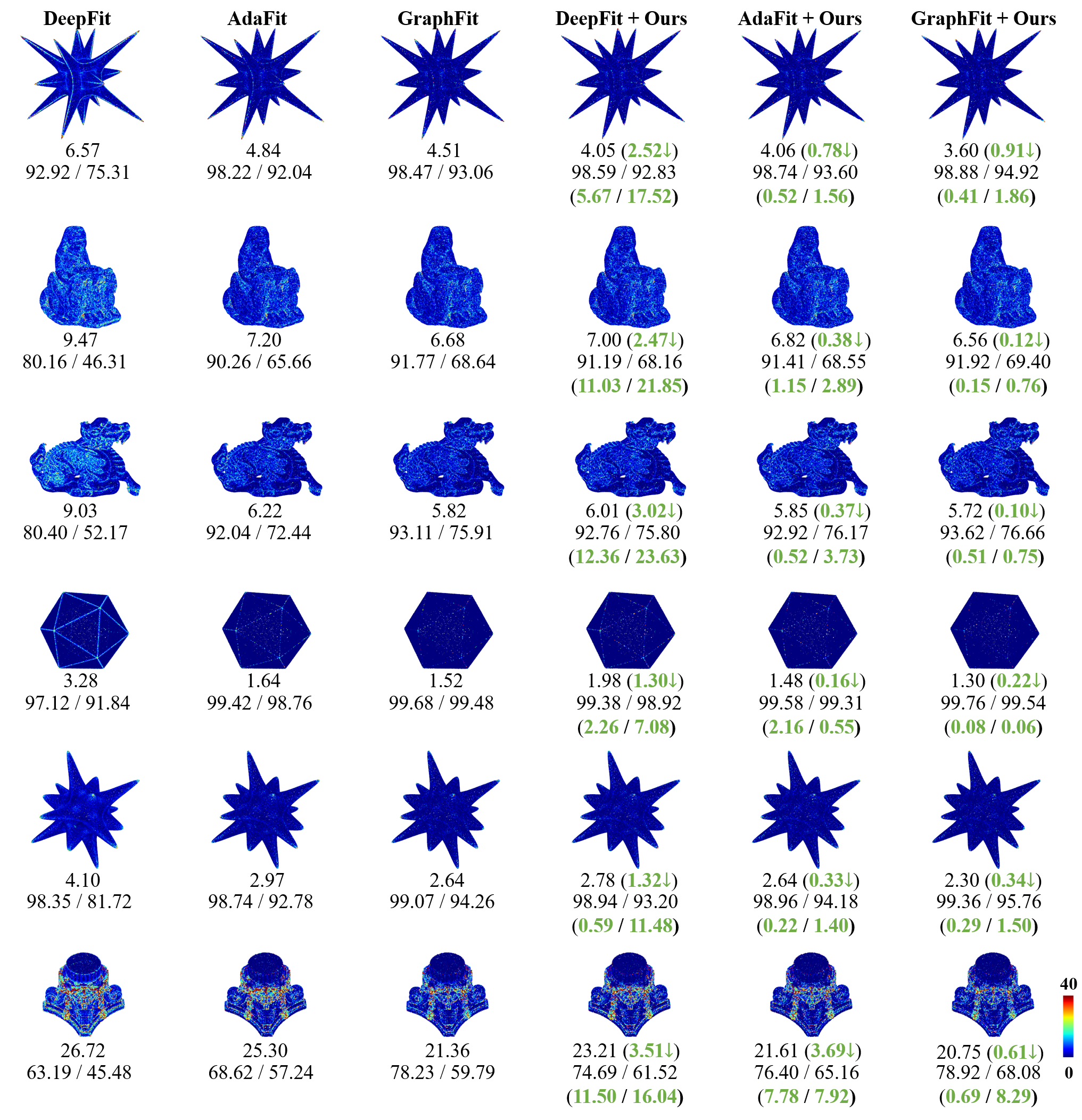}
    \caption{The error heatmap of normal estimation on PCPNet dataset. The first line of bottom values represents the average error, and second line reports the PGP10 and PGP5. The green values in  brackets denote the relative improvement on the corresponding baseline models. 
    The angle errors are mapped to a range from $0^\circ$ to $40^\circ$. }
    \label{fig_pcp_s}
\end{figure*}


\end{document}